%% file: main_arxiv.tex
\documentclass{article}

\usepackage{nac_preprint}

\usepackage{microtype}
\usepackage{graphicx}
\usepackage{subfigure}
\usepackage{booktabs} 

\usepackage{hyperref}

\usepackage{amsmath}
\usepackage{amssymb}
\usepackage{mathtools}
\usepackage{amsthm}
\usepackage{arydshln}

\usepackage{rotating}

\usepackage{algorithm}
\usepackage{algorithmicx}
\usepackage[noend]{algpseudocode}
\algnewcommand{\LineComment}[1]{\State \(//\) #1}

\usepackage{enumitem}

\usepackage{etoolbox}
\usepackage{tikz}
\usetikzlibrary{tikzmark}
\usetikzlibrary{calc}

\errorcontextlines\maxdimen

\newcommand{\ALGtikzmarkcolor}{black}
\newcommand{\ALGtikzmarkextraindent}{4pt}
\newcommand{\ALGtikzmarkverticaloffsetstart}{-.5ex}
\newcommand{\ALGtikzmarkverticaloffsetend}{-.5ex}
\makeatletter
\newcounter{ALG@tikzmark@tempcnta}

\newcommand\ALG@tikzmark@start{%
    \global\let\ALG@tikzmark@last\ALG@tikzmark@starttext%
    \expandafter\edef\csname ALG@tikzmark@\theALG@nested\endcsname{\theALG@tikzmark@tempcnta}%
    \tikzmark{ALG@tikzmark@start@\csname ALG@tikzmark@\theALG@nested\endcsname}%
    \addtocounter{ALG@tikzmark@tempcnta}{1}%
}

\def\ALG@tikzmark@starttext{start}
\newcommand\ALG@tikzmark@end{%
    \ifx\ALG@tikzmark@last\ALG@tikzmark@starttext
    \else
        \tikzmark{ALG@tikzmark@end@\csname ALG@tikzmark@\theALG@nested\endcsname}%
        \tikz[overlay,remember picture] \draw[\ALGtikzmarkcolor] let \p{S}=($(pic cs:ALG@tikzmark@start@\csname ALG@tikzmark@\theALG@nested\endcsname)+(\ALGtikzmarkextraindent,\ALGtikzmarkverticaloffsetstart)$), \p{E}=($(pic cs:ALG@tikzmark@end@\csname ALG@tikzmark@\theALG@nested\endcsname)+(\ALGtikzmarkextraindent,\ALGtikzmarkverticaloffsetend)$) in (\x{S},\y{S})--(\x{S},\y{E});%
    \fi
    \gdef\ALG@tikzmark@last{end}%
}

\apptocmd{\ALG@beginblock}{\ALG@tikzmark@start}{}{\errmessage{failed to patch}}
\pretocmd{\ALG@endblock}{\ALG@tikzmark@end}{}{\errmessage{failed to patch}}
\makeatother

\usepackage[capitalize,noabbrev]{cleveref}

\theoremstyle{plain}

\theoremstyle{definition}

\theoremstyle{remark}

\usepackage[textsize=tiny]{todonotes}

\title{Lifelong Neural Predictive Coding: Learning\\Cumulatively Online without Forgetting}

\author{
Alexander G. Ororbia\\
Rochester Institute of Technology\\
Rochester, NY 14623, USA\\
\texttt{ago@cs.rit.edu}
\And
Ankur Mali\\
The Pennsylvania State University\\
State College, PA 16801, USA\\
\texttt{ago@cs.rit.edu}
\AND
Daniel Kifer\\
The Pennsylvania State University\\
State College, PA 16801, USA\\
\texttt{ago@cs.rit.edu}
\And
C. Lee Giles\\
The Pennsylvania State University\\
State College, PA 16801, USA\\
\texttt{ago@cs.rit.edu}
}

\begin{document}

\setlength{\abovedisplayskip}{5pt}
\setlength{\belowdisplayskip}{5pt}
\setlength{\abovedisplayshortskip}{3pt}
\setlength{\belowdisplayshortskip}{3pt}




\maketitle

\begin{abstract}
In lifelong learning systems based on artificial neural networks, one of the biggest obstacles is the inability to retain old knowledge as new information is encountered. This phenomenon is known as catastrophic forgetting. In this paper, we propose a new kind of connectionist architecture, the Sequential Neural Coding Network, that is robust to forgetting when learning from streams of data points and, unlike networks of today, does not learn via the popular back-propagation of errors. Grounded in the neurocognitive theory of predictive processing, our model adapts synapses in a biologically-plausible fashion while another neural system learns to direct and control this cortex-like structure,  mimicking some of the task-executive control functionality of the basal ganglia. In our experiments, we demonstrate that our self-organizing system experiences significantly less forgetting compared to standard neural models, outperforming a swath of previously proposed methods, including rehearsal/data buffer-based methods, on both standard (SplitMNIST, Split Fashion MNIST, etc.) and custom benchmarks even though it is trained in a stream-like fashion. Our work offers evidence that emulating mechanisms in real neuronal systems, e.g., local learning, lateral competition, can yield new directions and possibilities for tackling the grand challenge of lifelong machine learning.
\end{abstract}

\section{Introduction}
\label{sec:intro}

Lifelong learning is a part of machine learning and artificial intelligence research with the goal of developing a computational agent that can learn over time and continually accumulate new knowledge while retaining its previously learned experiences \cite{thrun1995lifelong,parisi2018continual}. For example, suppose an agent needs to learn to classify digits, then types of clothing, and then sketches of objects. As each new task arrives the agent is expected to process the accompanying data and learn the new task but still remember how to complete the old tasks, at least without significant degradation in performance or loss of generalization.
Modern-day connectionist systems are typically trained on a fixed pool of data samples, collected in controlled environments, in isolation and random order, and then evaluated on a separate validation data pool. This is a far cry from what we really desire from learning machines. 

When we look to humans or other animals, we see that they are more than capable of learning in a continual manner, making decisions based on sensorimotor input throughout their lifespans \cite{power2017neural}. This ability to incrementally acquire and refine knowledge over long periods of time is driven by cognitive processes that come together to create the experience-driven specialization of motor and perceptual skills \cite{zenke2017hebbian,power2017neural}. Thus, evaluating how neural systems generalize on task sequences, as opposed to single, isolated tasks, proves to be a far greater challenge. 
In order to continually adapt, the brain must retain specific memories of prior tasks.  In working towards the challenge of lifelong machine learning, this paper makes the following contributions: 
1) we propose the sequential neural coding network, an interactive generative model that jointly reconstructs input and predicts its label, and an algorithm for updating its synapses, 
2) we show that memory retention is vastly improved by integrating our model's multi-step nature with lateral competition that is driven by a task selection function inspired by the basal ganglia \cite{leisman2014cognitive}, and 
3) we compare our model's performance against state-of-the-art baselines, including both regularization and rehearsal/replay-based methods, on publicly-available benchmarks.

\section{Related Work}
\label{sec:rw}

It is well-known that when artificial neural networks (ANNs) are trained on more than one task sequentially, the new information contained in subsequent tasks leads to catastrophic interference (a.k.a. catastrophic forgetting) with the information acquired in earlier tasks \cite{mccloskey_catastrophic_1989,ratcliff_connectionist_1990,lewandowsky1994relation}. This happens in connectionist systems when the new data instances to be learned are significantly different from previously observed ones. This causes the new information to overwrite knowledge currently encoded in the system's synaptic weights, due to the sharing of neural representations over tasks \cite{french1999catastrophic,mccloskey_catastrophic_1989} (this is known as the representational overlap problem). In isolated, task-specific (offline) learning (though it still occurs \cite{toneva2018empirical}), this type of weight overwriting occurs to a lesser degree because the patterns are presented to the agent pseudo-randomly and multiple times, i.e., via multiple epochs.

Over the decades, there have been many approaches proposed to mitigate catastrophic forgetting in neural systems. Some of the earliest attempts proposed memory systems where prior data points were stored and regularly replayed to the network in a process called ``rehearsal'', which involved interleaving these data points with samples drawn from new datasets \cite{robins1993catastrophic,robins1995catastrophic,robins1996consolidation,gepperth2016bio,rebuffi2017icarl}. Though effective, the main drawback of these approaches is that they require explicitly storing old data. Such a mechanism is not known to exist in the brain and, as a practical matter, this leads to exploding working (hardware) memory requirements (inefficiency). In addition, rehearsal-based approaches do not offer any mechanisms to preserve consolidated knowledge in the face of acquiring new information \cite{zenke2017hebbian}.
Other approaches attempt to allocate additional neural ``resources'', i.e., growing the networks when required \cite{rusu2016progressive,parisi2017lifelong}, motivated by earlier findings \cite{meier1991neurotransmitters}. However, this leads to dramatically increasing computational requirements over time as the networks grow larger. To compound these issues further, systems with growing capacity cannot know how many resources to allocate at a given time since the number of tasks and samples are not known a priori (without imposing strong assumptions on the input distribution). Other approaches try to block old information from being overwritten through regularization \cite{kirkpatrick2017overcoming}. From this vast collection of research, each approach bearing strengths and weaknesses, three suggested remedies have emerged: 1) allocate additional neural resources to accommodate new knowledge, 2) use non-overlapping representations (or semi-distributed ones \cite{french1992semi}) if resources are fixed, and 3) interleave old patterns with new ones as new information is  acquired.

In this work, we consider the setting where the space available to the agent \textcolor{black}{grows slowly compared to the rate} of new tasks being presented. This means that we cannot just create a new, separate network for every task observed in the stream. Furthermore, storing, reshuffling, and re-presenting the data in the stream is not feasible in this setting. \textcolor{black}{Concretely, our approach could be classified as class incremental learning (Class-IL) \cite{van2019three} given that we \textbf{do not use task descriptors/identifiers} at both training and test time.}
In addition to addressing the above, our contribution to lifelong learning is motivated by premise that developing algorithms  \cite{movellan1991contrastive,scellier2017equilibrium,lee2015difference,ororbia2018biologically,bartunov2018assessing} that serve as alternatives to back-propagation 
will lead to the creation of promising architectures with mechanisms better equipped to tackle problems like catastrophic interference. 

\section{Cumulative Learning with Neural Coding}
\label{sec:lml_problem}

\paragraph{Notation: }
\label{sec:notation}
We start by defining key notation.
$\odot$ indicates a Hadamard product while $\cdot$ denotes matrix/vector multiplication. $(\mathbf{v})^T$ is the transpose of $\mathbf{v}$. Matrices/vectors are depicted in bold-font, e.g., matrix $\mathbf{M}$ or vector $\mathbf{v}$ (scalars shown in italic). Finally, $||\mathbf{v}||_2$ denotes the Euclidean norm of $\mathbf{v}$.

\subsection{Sequential Learning and the Data Continuum}
\label{sec:problem_setup}

This work focuses on learning a neural system in the context of sequential learning. Starting from an early definition \cite{thrun1996learning} of this form of learning, we assume there is a sequence of tasks $\mathcal{T}_{1}, \mathcal{T}_{2}, \mathcal{T}_{3}, \dots$ (with potentially no end to the sequence) that are presented to a system in order. When faced with the $(N+1)$th task, the system should use the knowledge it has gained from the previous $N$ tasks to aid in learning and performing the current task.
The knowledge of a system is stored in a knowledge base (KB), e.g., the synapses of a neural model.  Each task $\mathcal{T}_i$ has its own corresponding dataset $D_i=\{(\mathbf{y}_1,\mathbf{x}_1,t_i) \dots (\mathbf{y}_{n_i},\mathbf{x}_{n_i},t_i)\}$ with $n_i$ examples. Here $\mathbf{x}_j \in \mathcal{R}^{J_x \times 1}$ represents the feature vector of the $j^\text{th}$ example ($J_x$ is its dimensionality), $\mathbf{y}_j \in \{0,1\}^{J_y \times 1}$ is the target (label), and $t_i$ is the task descriptor 
that identifies $(\mathbf{y}_j, \mathbf{x}_j)$ as being a data point from task $i$. 
The task descriptor is one-hot encoded as $\mathbf{t}  \in \{0,1\}^{(N+1) \times 1}$ and, when a new task is encountered, the size of the one-hot vector is increased by one -- thus the network adds an extra randomly initialized input node if it accepts $t_i$ as an extra input. Note that while we present the task data continuum with $t_i$ explicitly depicted, we will generalize our models to not depend on $t_i$ (making this problem task descriptor-free).

\noindent
\textbf{Dynamic Output Units:} 
The output nodes in our setting get re-used for each new task. For example, output node $1$ of the network could represent a prediction for the digit ``1" in task $\mathcal{T}_1$ (e.g., digit recognition), while in task $\mathcal{T}_2$ (e.g., clothing recognition) the same node could represent a prediction for ``pants''. If a new task has more classes than previous tasks, we add output nodes with randomly initialized weights. For example, if prior tasks were binary and the new task has $4$ targets, we add $2$ more outputs to the network. This is a difficult form of cumulative learning \cite{fei2016learning}.

\noindent
\textbf{Context Units:}
For the model that we develop in this study, when task $t$ is encountered, for each layer $\ell$ in the network, we make use of a task embedding vector $\mathbf{g}^\ell_t$ (this new memory is much smaller than creating a new network for task $t$, which would require new weight matrices per layer rather than an extra vector). 
All $\mathbf{g}^\ell_t \in \mathcal{R}^{J_\ell \times 1}$ ($J_\ell$ is the number of units in layer $\ell$) are stored in a memory matrix $\mathbf{M}^\ell  \in \mathcal{R}^{(N+1) \times J_\ell}$ \textcolor{black}{($\mathcal{M} = \{\mathbf{M}^1,...\mathbf{M}^L\}$)}, where a context can be retrieved using a one-hot encoding of the task descriptor, i.e., $\mathbf{M}^\ell \cdot \mathbf{t}$ \textcolor{black}{( this will be produced by another system -- see Section \ref{sec:basal_ganglia}}).

\subsection{The Interactive Generative Model}
\label{sec:generative_model}

The sequential neural coding network (S-NCN) is designed to make flexible conditional predictions -- e.g., predicting $\mathbf{y}$ given $\mathbf{x}$, predicting both $\mathbf{y}$ and $\mathbf{x}$, predicting the missing parts of $\mathbf{x}$ given $\mathbf{y}$ and the observed parts of $\mathbf{x}$, etc. In order to do so, it treats inputs and outputs in a non-standard way. The input to the model is a task descriptor $t_i$ and the output units represent $(\mathbf{y}, \mathbf{x})$. To predict $\mathbf{y}_i$ given $\mathbf{x}_i$, we clamp the output nodes that are responsible for predicting $\mathbf{x}$ and force their output to be $\mathbf{x}_i$.\footnote{Similarly, in the case of missing data, we can ask the network to predict the $\mathbf{y}_i$ and the missing parts of $\mathbf{x}_i$ given the observed parts of $\mathbf{x}_i$ by clamping outputs to only the observed parts of $\mathbf{x}_i$.} During training, outputs are clamped to both $\mathbf{y}_i$ and $\mathbf{x}_i$, forcing the S-NCN to update latent states and synapses. 
The S-NCN can operate as a probabilistic generative network by feeding in a random noise vector as input, but we leave this extension to future work (our focus here is predicting $\mathbf{y}$ given $\mathbf{x}$).

The full computational process of the S-NCN is defined by three key steps: 
1) layer-wise hypothesis generation, 
2) state error-correction, and 
3) synaptic weight evolution. 
The first two steps iteratively predict and correct the representations of the model for the input and target values of the task. After $K$ iterations, model weights and the current task descriptor memory are updated.
In this section, we provide details of the above steps and then describe the objective function that our model optimizes.

\subsubsection{Inference: Predicting and Correcting States} 
\label{subsec:prediction}

\textbf{Layer-wise State Prediction.}
The architecture of the S-NCN can be viewed as a stack of parallel, stateful neural-based prediction layers $P_1,...,P_\ell,..., P_m$, where the goal of each predictor is to guess the internal state of the predictor in the layer below (i.e., the S-NCN is not a feedforward network).
The state of $P_\ell$ (layer $\ell$) is represented by the (zero-initialized) vector $\mathbf{z}^\ell  \in \mathcal{R}^{J_\ell \times 1}$ ($J_\ell$ is the number of units in layer $\ell$). $P_\ell$ makes a prediction $\mathbf{z}_\mu^{\ell-1}$ about the current state $\mathbf{z}^{\ell-1}$ of $P_{\ell-1}$ (layer $\ell-1$). Furthermore, we let $\mathbf{z}^0_{x}$ and $\mathbf{z}^0_{y}$ denote clamped outputs. That is, if we want to predict the label $\mathbf{y}_i$ given the features $\mathbf{x}_i$, we set $\mathbf{z}^0_x=\mathbf{x}_i$ and if we want to train, we set both $\mathbf{z}^0_x=\mathbf{x}_i$ and $\mathbf{z}^0_y=\mathbf{y}_i$. The values predicted by layer $1$ are denoted by $\mathbf{z}^0_{\mu, x}$ and $\mathbf{z}^0_{\mu, y}$. Note that $\mathbf{z}^0 = [ \mathbf{z}^0_x, \mathbf{z}^0_y ]$, where the two are concatenated (if either is missing, the SNCN completes the required values -- see Appendix).

With respect to neural structure, the parallel predictors of the S-NCN are locally connected through (forward) generative weights $\mathbf{W}^\ell \in \mathcal{R}^{J_\ell \times J_{\ell+1}}$ and error (feedback) weights $\mathbf{E}^\ell \in \mathcal{R}^{J_{\ell+1} \times J_\ell}$, which work to transmit error information to relevant regions of neural processing elements, effectively coordinating all of the model's predictors. 
Formally, a predictor, with state $\mathbf{z}^{\ell+1}$, that guesses the state of $\mathbf{z}^\ell$, takes on the form (given matrix $\mathbf{W}^{\ell+1}$ and activation $\phi^{\ell+1}$):
\begin{align}
    \mathbf{z}^{\ell}_\mu = \mathbf{W}^{\ell+1} \cdot  \phi^{\ell+1}( \mathbf{z}^{\ell+1} ),\quad \mathbf{e}^\ell = (\mathbf{z}^{\ell} - \mathbf{z}^{\ell}_\mu) \label{eqn:predict_error}
\end{align}
where $\mathbf{e}^\ell$ is a block of error units. Error units are paired with each predictor. Their task is to compute the disagreement or mismatch between the predictor's output and the target activity pattern $\mathbf{z}^{\ell}$. The error unit vector $\mathbf{e}^\ell$ can also be derived from the total discrepancy reduction presented in Equation \ref{eqn:total_discrep}. Note that for layer $0$, $\mathbf{e}^0 = [\mathbf{e}^0_x,\mathbf{e}^0_y]$ (there are error neurons for $\mathbf{x}_i$ as well as others for  $\mathbf{y}_i$, the concatenation of which makes up the bottom-most prediction error signal).

\textbf{Latent State Correction and Context Updating.} Once each layer $\ell$ has made a prediction about the layer below it and error units have been activated, the state of layer $\ell$ can be adjusted to take into account the local top-down and bottom-up error information. Using its current state and the error nodes (along with its task embedding vector $\mathbf{g}^\ell_t$), layer $\ell$ in the S-NCN adjusts its state $\mathbf{z}^\ell$ as follows:
\begin{align}
    \mathbf{z}^\ell(k) &= f^\ell( \mathbf{z}^\ell(k-1) + \beta \mathbf{d}^\ell, \mathbf{g}_t^\ell ),\quad \text{where } \mathbf{d}^\ell = -\mathbf{e}^\ell + \mathbf{E}^\ell \cdot \mathbf{e}^{\ell-1}  \label{eqn:state_correct}
\end{align}
where $\beta$ is the state correction rate and $k$ marks one step of the S-NCN's $K$-step inference process.
Note that $\mathbf{d}^\ell$ is the perturbation that adjusts the values of the hidden states -- it combines a top-down expectation of layer $\ell$ w/ a bottom-up pressure from layer $\ell-1$).
The error feedback weights $\mathbf{E}^\ell$ are parameters that play a crucial role in this calculation, as they are responsible for transmitting the error for $\ell$ back up to layer $\ell+1$. Notably, part of the state-correction requires competition among the individual units in a given layer through the function $f^\ell(\mathbf{z}^\ell, \mathbf{g}^\ell_t)$. There are various ways in which this function can be implemented and, in this study, we implemented it as:
\begin{align*}
    f^\ell(\mathbf{z}^\ell, \mathbf{g}_t^\ell) = \big( \mathbf{I} \odot \mathbf{V}^\ell \big) \cdot \mathbf{z}^\ell, \mbox{where, } \mathbf{V}^\ell = \Call{bkWTA}{\mathbf{g}^\ell_t,K} \cdot \Call{bkWTA}{(\mathbf{g}^\ell_t)^T,K}
\end{align*}
where $\Call{bkWTA}{\mathbf{v},K}$ is the binarized $K$ winners-take-all function, yielding a binary vector with $1$ at the index of each of the $K$ winning units, or formally:
\begin{align*}
\Call{bkWTA}{\mathbf{v},K} = \big\{ 1 \mbox{ if } v_j \in \{K \mbox{ largest elements of } \mathbf{v}\} \mbox{ and } \phi(\mathbf{v}) = 0 \mbox{ otherwise} \big\} \mbox{.}
\end{align*}
In the Appendix, we study other forms of the competition function (and find that the above performed best, so we report this in the main paper). Note that this lateral inhibition is a function of evolving context $\mathbf{g}^\ell$, triggered by a task pointer $t_i$ (produced by a task selector model, which we define later).

In real neural systems, competition between units within a layer is thought to facilitate contextual processing \cite{adesnik2010lateral}, where only some neuronal signals are strengthened while the activity of others is suppressed. Moreover, lateral competition, often modeled in classical models with anti-Hebbian learning \cite{foldiak1990forming}, encourages the formation of sparse codes \cite{olshausen1997sparse,szlam2011structured}. Since the S-NCN is an interactive/iterative model, incorporating lateral competition (in the form of a task-driven memory matrix) is natural and computed online, highlighting the model's flexibility compared to an ANN -- to incorporate lateral synaptic activity, recurrent weights could be introduced to an ANN but training it would requiring unrolling via backprop through time. More importantly, a lateral mechanism in the S-NCN works as an inductive bias that leads it to acquire task-dependent representations, which can aid in memory retention since the system has to store information on multiple, disjoint tasks. In a sense, this task specialization that we build into the neural dynamics is similar in spirit to activation sharpening \cite{french1991using}. 


\begin{figure*}[!t]
\begin{minipage}[t]{.5\textwidth}
\vspace{0pt}  
\begin{algorithm}[H]
\caption{State inference procedure.}
\label{alg:inference}
\fontsize{8.5}{9}\selectfont
\begin{algorithmic}[1]
   \State {\bfseries Input:} sample $(\mathbf{y},\mathbf{x}, \mathbf{t})$, $\beta$\textcolor{black}{, $\mathcal{M}$}, \& $\Theta$
   \Function{InferStates}{$(\mathbf{y},\mathbf{x}, \mathbf{t}), \Theta$}
   		\State $(\mathbf{g}^1,...,\mathbf{g}^\ell,...,\mathbf{g}^L) \leftarrow getContexts(\mathbf{t}\textcolor{black}{, \mathcal{M}})$
   		\State Set $\mathbf{z}^1,...,\mathbf{z}^\ell,...,\mathbf{z}^L$ to $\mathbf{0}$
   		\State $\mathbf{e}^L =\mathbf{0}$, $\mathbf{z}^0_x = \mathbf{x}$, $\mathbf{z}^0_y = \mathbf{y}$
   		\For{$k = 1$ to $K$}
   		\For{$\ell = L$ to $1$} \Comment{Run layerwise predictors}
   		\If{$\ell > 1$} \Comment{Latent prediction layer}
    	    \State $\mathbf{z}^{\ell-1}_\mu = \mathbf{W}^\ell \cdot  \phi^\ell( \mathbf{z}^\ell )$
    	    \State $\mathbf{e}^{\ell-1} = (\mathbf{z}^{\ell-1} - \mathbf{z}^{\ell-1}_\mu)$
    	\Else \Comment{Sensory prediction layer}
    	    \State $\mathbf{z}^{\ell-1}_{\mu,x} = \mathbf{W}^\ell_x \cdot  \phi^\ell( \mathbf{z}^\ell )$
    	    \State $\mathbf{e}^{\ell-1}_x = (\mathbf{z}^{\ell-1}_x - \mathbf{z}^{\ell-1}_{\mu,x})$
    	    \State $\mathbf{z}^{\ell-1}_{\mu,y} = \mathbf{W}^\ell_y \cdot  \phi^\ell( \mathbf{z}^\ell )$
    	    \State $\mathbf{e}^{\ell-1}_y = (\mathbf{z}^{\ell-1}_y - \mathbf{z}^{\ell-1}_{\mu,y})$
    	\EndIf
    	\EndFor
    	\For{$\ell = L$ to $1$} \Comment{Correct internal states}
    	\If{$\ell$ == $1$}
    	    \State $\mathbf{d}^\ell = -\mathbf{e} + \mathbf{E}^\ell_x \cdot \mathbf{e}^{\ell-1}_x + \mathbf{E}^\ell_y \cdot \mathbf{e}^{\ell-1}_y$
    	\Else
    	    \State $\mathbf{d}^\ell = -\mathbf{e} + \mathbf{E}^\ell \cdot \mathbf{e}^{\ell-1}$
    	\EndIf
    	\State $\mathbf{z}^\ell = f^\ell\Big( \phi^\ell( \mathbf{z}^\ell ) + \beta \mathbf{d}^\ell , \mathbf{g}^\ell \Big)$
    	\EndFor
    	\EndFor
        \State $\Lambda = (\mathbf{z}^1,...,\mathbf{z}^\ell,...,\mathbf{z}^L,\mathbf{d}^1,...,\mathbf{d}^\ell,...,\mathbf{d}^L,$
        \State \hspace{0.775cm} $\mathbf{e}^0,...,\mathbf{e}^\ell,...,\mathbf{e}^{L-1} )$
        \State \textbf{Return} $\Lambda$
    \EndFunction
\end{algorithmic}
\end{algorithm}
\end{minipage}
\begin{minipage}[t]{.495\textwidth}
\vspace{0pt}  
\begin{algorithm}[H]
\caption{Weight update computation.}
\label{alg:update}
\fontsize{8.5}{9}\selectfont
\begin{algorithmic}[1]
   \State {\bfseries Input:} $\Lambda$, $\lambda$, $\gamma$\textcolor{black}{, $\mathcal{M}$}, \& $\Theta$
   \Function{UpdateWeights}{$\Lambda, \Theta$}
        \LineComment Calculate weight displacements
        \For{$\ell = L$ to $1$}
            \State $\Delta \mathbf{W}^\ell = \big( \mathbf{e}^{\ell-1} \cdot (\phi^\ell( \mathbf{z}^\ell ))^T\big) \odot \mathbf{S}^\ell_W$ 
            \State $\Delta \mathbf{E}^\ell = \gamma ( \mathbf{d}^\ell \cdot (\mathbf{e}^{\ell-1})^T ) \odot \mathbf{S}^\ell_E$
        \EndFor
   		\State $\Delta \mathbf{W}^1_x = \big( \mathbf{e}^0_x \cdot (\phi^1( \mathbf{z}^1 ))^T \big) \odot \mathbf{S}^1_{W,x} $
   		\State $\Delta \mathbf{E}^1_x = \gamma ( \mathbf{d}^1 \cdot (\mathbf{e}^0_x)^T )  \odot \mathbf{S}^1_{E,x} $
   		\State $\Delta \mathbf{W}^1_y = \big( \mathbf{e}^0_y \cdot (\phi^1( \mathbf{z}^1 ))^T \big) \odot \mathbf{S}^1_{W,y} $
   		\State $\Delta \mathbf{E}^1_y = \gamma ( \mathbf{d}^1 \cdot (\mathbf{e}^0_y)^T )  \odot \mathbf{S}^1_{E,y} $
   	    \State 
   		\LineComment Update current weights
   		\For{$\ell = L$ to $1$}
   		    \State $\mathbf{W}^\ell = \mathbf{W}^\ell + \lambda \Delta \mathbf{W}^\ell$
   		    \State $\mathbf{E}^\ell = \mathbf{E}^\ell + \lambda \Delta \mathbf{E}^\ell$
   		\EndFor
   		\State $\mathbf{W}^1_x = \mathbf{W}^1_x + \lambda \Delta \mathbf{W}^1_x$
   		\State $\mathbf{E}^1_x = \mathbf{E}^1_x + \lambda \Delta \mathbf{E}^1_x$
   		\State $\mathbf{W}^1_y = \mathbf{W}^1_y + \lambda \Delta \mathbf{W}^1_y$
   		\State $\mathbf{E}^1_y = \mathbf{E}^1_y + \lambda \Delta \mathbf{E}^1_y$
   		\State Update  $(\mathbf{g}^1,\cdots,\mathbf{g}^L, \textcolor{black}{\mathcal{M}})$ \textcolor{black}{via Eqn. 4} 
   		\LineComment Return new weights
   		\State $\Theta = \{\mathbf{W}^1_x,\mathbf{W}^1_y,...,\mathbf{W}^\ell...\mathbf{W}^L,$
   		\State \hspace{0.75cm} $\mathbf{E}^1_x,\mathbf{E}^1_y,...,\mathbf{E}^\ell,...,\mathbf{E}^L \}$
   		\State \textbf{Return} $\Theta$
   		\vspace{0.075cm}
   	\EndFunction
\end{algorithmic}
\end{algorithm}
\end{minipage}
\vspace{-0.55cm}
\end{figure*}

\subsubsection{Updating Synaptic Parameters}
\label{subsec:update}

Given that the S-NCN is an interactive network \cite{mcclelland1981interactive}, inferring its states requires running Equations \ref{eqn:predict_error} and \ref{eqn:state_correct} $K$ times, where the model alternates between making predictions and then correcting states once error units have been computed. Once latent states have been inferred, the S-NCN is then able to adjust its synaptic values.
The synaptic updates take the form of Hebbian-like rules:
\begin{align}
    \Delta \mathbf{W}^\ell = \mathbf{e}^\ell \cdot (\phi^\ell( \mathbf{z}^{\ell+1} ))^T \odot \mathbf{S}^\ell_W, \; \mbox{and, } \; 
    \Delta \mathbf{E}^\ell = \alpha \Big( \mathbf{d}^{\ell+1} \cdot ( \mathbf{e}^\ell )^T \Big) \odot \mathbf{S}^\ell_E
\end{align}
where $\alpha$ is a scaling factor, usually set to $< 1.0$, that makes the error feedback weights change at a slower rate than the forward weights (this improves convergence \cite{ororbia2018continual}). $\mathbf{S}^\ell_W  \in \mathcal{R}^{J_\ell \times J_{\ell+1}}$ and $\mathbf{S}^\ell_E  \in \mathcal{R}^{J_{\ell+1} \times J_\ell}$ are modulation factors that provide stability to the weight updates (see Appendix). 

An important property of the above weight update rules is that they are local -- to compute changes in the synapses, all we require is the information immediately available to the neuron(s) of interest (making these rules function similarly to classical Hebbian updates \cite{hebb1949organization}, although there are important differences to them, as discussed in \cite{ororbia2018continual}). Since a neuron is able to immediately generate a hypothesis given its own internal state, without requiring the active generation of other predictors, and its error can be readily computed after prediction by comparing to the current state of the target neuron state, the weight updates of any predictor layer may be computed in parallel to others. This would allow us to allocate dedicated computing cores to particular predictors, or ``pieces'', of the S-NCN. 
Observe that the S-NCN does not require activation derivatives in any of its computations (this neurobiologically more realistic and favorable for specialized hardware implementations). 

Furthermore, during the learning process, each context vector $\mathbf{g}_t^\ell$ and the corresponding memory matrix is adjusted according to the following simple contrastive rule:
\begin{align}
    \textcolor{black}{\mathbf{g}^\ell_{t+1}} = \mathbf{g}^\ell_t + \eta_e \mathbf{d}^\ell - \eta_g ( \mathbf{g}^\ell_t - \frac{1}{t-1} \sum^{t-1}_{j = 1}(\mathbf{g}^\ell_j) ), \; \mbox{and, } \Delta \mathbf{M}^\ell &= \mathbf{g}^\ell_{t+1} *  (\mathbf{t}_j)^T  \label{eqn:code_update}
\end{align}
where $\eta_e$ modulates a long-term memory update using the current perturbation to be applied to layer $\ell$. $\eta_g$ controls the repulsion term, which ``pushes'' context codes away from each other (for diversity). These adapted codes, which influence inter-neuronal competition in a task-sensitive manner, could be viewed as a simplification of distributed temporal context \cite{howard2002distributed}, where contiguity, i.e., recall/generation of one item is influenced by the presence of another, is introduced into S-NCN distributed processing.

The pseudocode illustrating how the elements described so far are combined in an S-NCN system is presented in Algorithms \ref{alg:inference} and \ref{alg:update}.
The transmission of bottom-up and top-down errors in the S-NCN is motivated by the theory of predictive processing \cite{rao1997dynamic,rao1999predictive,huang2011predictive,wacongne2012neuronal,chalasani2013deep,clark2015surfing,santana2017exploiting} 
and classical work on interactive networks \cite{mcclelland1981interactive,mcclelland1993grain,o1996biologically}, where  models undergo a settling process to process input stimuli more than once (see Appendix for a discussion on motivations). Though this requires extra computation, the process endows the network with desirable properties, e.g., the ability to conduct constraint satisfaction \cite{oreilly1998sixprinciples,oreilly2001inhibition}.
By using a multi-step processing, laterally-competitive processing scheme, the S-NCN is able to ``select'' subnetworks (portions of neurons) for specific tasks, reducing representational overlap and, ultimately, forgetting. This selection is driven by the task pointer, produced by the task selector, $\mathcal{T}(\mathbf{x})$, the final piece of the S-NCN system, which we describe next.

\subsubsection{Objective Function}
\label{subsec:objective}

During training, when presented with stimulus $(\mathbf{y}_i,\mathbf{x}_i,\mathbf{t}_i)$, the S-NCN adjusts its internal states and synapses so that the output $\mathbf{z}^0_\mu$ of layer $1$ ($P_1$) is as close as possible to  $(\mathbf{y}_i,\mathbf{x}_i)$. It does this by minimizing \emph{total discrepancy} \cite{ororbia2017learning} --  a measure of its total internal disorder (and one that can be shown to approximate free energy \cite{friston2009free}), which is the sum of all mismatches between layerwise guesses and actual states. In its general form, total discrepancy for an S-NCN is formally: 
\begin{align}
    \mathcal{L}(\Theta) = \sum^{L-1}_{\ell=0} \mathcal{L}^\ell(\Theta^\ell) \mbox{, where } \mathcal{L}^\ell(\Theta^\ell) = \frac{1}{2} (|| \mathbf{z}^\ell - \mathbf{z}^\ell_{\mu} ||_2)^2
    \mbox{.}
    \label{eqn:total_discrep}
\end{align}
$\Theta$ contains all of the 
synaptic parameters, i.e., $\Theta = \{\mathbf{W}^{1},\mathbf{E}^{1},\cdots,\mathbf{W}^{L},\mathbf{E}^{L} \}$ and $\Theta^\ell = \{\mathbf{W}^{\ell},\mathbf{E}^{\ell}\}$.

The above loss decomposes the problem of credit assignment in the S-NCN into several sub-problems that each focuses on the comparison between the prediction $\mathbf{z}^\ell_\mu$ made by layer $\ell+1$ and the actual state value $\mathbf{z}^\ell$ of layer $\ell$. The resulting updates to each state $\mathbf{z}^\ell$, as well as the relevant parameters, will then depend on a bottom-up transmitted error signal and the top-down influence of the mismatch with the expectation of the predictor immediately above \cite{ororbia2017learning,ororbia2018continual}.
Note that, while we motivate aspects of our model from a neuro-cognitive perspective, the error units and weight updates 
can be derived from the total discrepancy function above \cite{ororbia2018continual} (and cast as approximately minimizing free energy).

\begin{figure}
     \centering
     \begin{subfigure}
         \centering
         \includegraphics[width=0.50\textwidth]{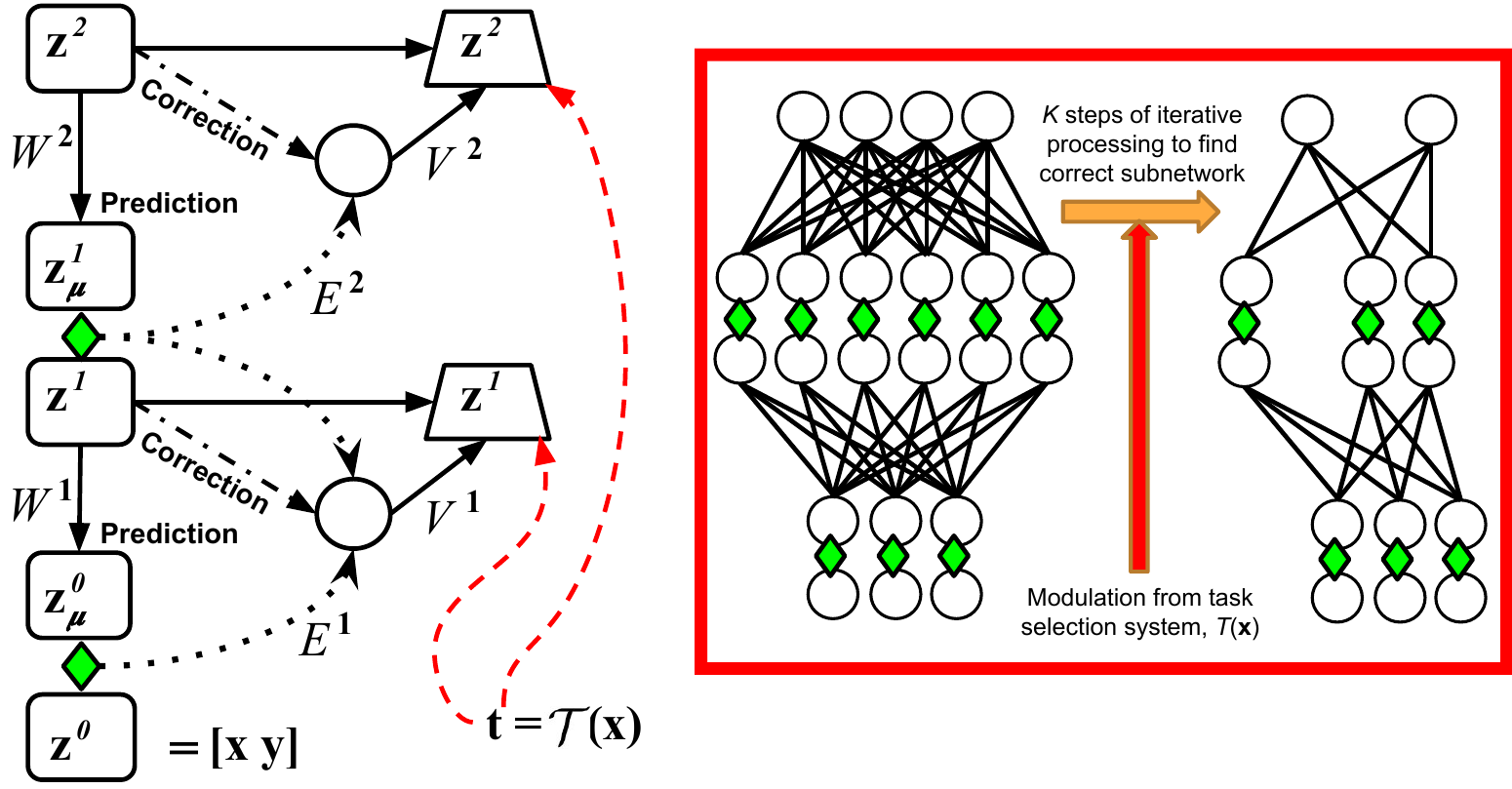}
     \end{subfigure}
     \begin{subfigure}
         \centering
         \includegraphics[width=0.375\textwidth]{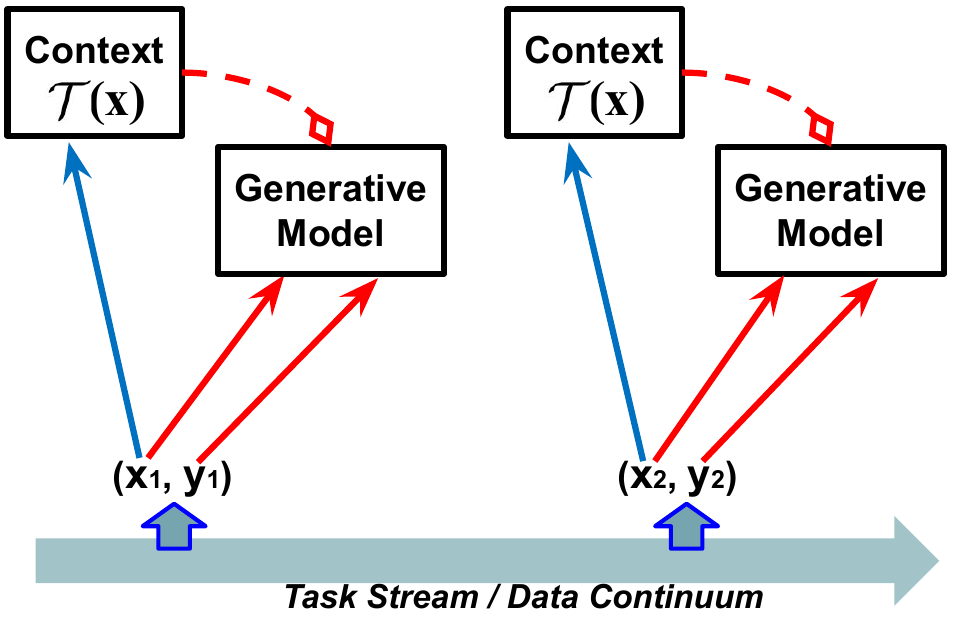}
     \end{subfigure}
     \vspace{-0.35cm}
\caption{(Left) The sequential neural coding network is shown processing an input $(\mathbf{y},\mathbf{x})$ over one step, given a task context $\mathbf{t}$ (produced by the task selection model $\mathcal{T}(\mathbf{x})$), overriding the current $\mathbf{z}^\ell$ after error-correction and application of lateral inhibition matrices $\{\mathbf{V}^1,\mathbf{V}^2\}$. Inside the red box is shown one possible emergent subnetwork. 
Green diamonds indicate error units. (Right) The full complementary neural system shown processing patterns from a data continuum.}
\label{fig:ncn}
\vspace{-0.55cm}
\end{figure}


\subsection{The Neural Task Selection Model} %
\label{sec:basal_ganglia}

In the last section, we described the S-NCN as taking in a task signal that drives the choice of the right internal task context memory $\mathbf{g}^\ell$ for layer $\ell$. To create this task pointer, we design a second neural model that 
we call the task selector $\mathcal{T}(\mathbf{x})$ -- this will allow
the S-NCN to automatically decide when a new task has been encountered and to determine, at test time, what task data points belong to. The motivation behind $\mathcal{T}(\mathbf{x})$ comes from the neuroscientific idea that the basal ganglia plays the role of information routing (among other roles), which serves as a form of task selection \cite{yehene2008basal,buschman2014goal}). In other words, it selects or enables various cognitive programs stored in other cortical regions \cite{leisman2014cognitive}. As a result, we develop a sort of complementary learning system (CLS) (different than that of \cite{mcclelland1995there}, which sets up an interaction between a hippocampus and a neocortex model), which pairs our cortex-like model with a basal ganglia-like model. We refer to our task selector $\mathcal{T}(\mathbf{x})$ as the ``functional neural basal ganglia'' (FNBG) in order to emphasize the fact that the actual basal ganglia in the human brain is far more complex and does far more than what our simple computational model does. 
Our FNBG model, built with competitive learning, has two roles:
1) task shift detection - deciding whether data from an input stream is indicating the presence of a new task, and 
2) task recognition - deciding whether incoming data requires switching to an existent task context or creating a new context knowledge.


\paragraph{Task Shift Detection: }
In order to detect the occurrence of a new task while processing data from the pattern stream, $\mathcal{T}(\mathbf{x})$ utilizes the output error neurons of the generative model described earlier to detect spikes in their values that might indicate distributional/context drift. Specifically, $\mathcal{T}(\mathbf{x})$ maintains an exponential running mean $\mu_{\mathcal{L}}$ and variance $\sigma^2_{\mathcal{L}}$ of the norm of the cortex model's label error neurons $\mathbf{e}^0_y$. The necessary statistics are calculated as follows:
\begin{align}
    \Delta &= ||\mathbf{e}^0_y||^2_2 - \mu_{\mathcal{L}}(t-1) \label{eqn:shift_delta}\\
    \mu_{\mathcal{L}}(t) &= \mu_{\mathcal{L}}(t-1) + \eta \Delta \label{eqn:shift_mean}\\
    \sigma^2_{\mathcal{L}}(t) &= (1 - \eta) \sigma^2_{\mathcal{L}}(t-1) + \eta (\Delta)^2 \label{eqn:shift_var}
\end{align}
with $\eta = 0.1$ (determined by preliminary experiments).
Using the above dynamic statistics, a shift is detected by determining if the following inequality evaluates to true (repeatedly for a series of $5$ consecutive batches): $\mu_{\mathcal{L}}(t) >  \mu_{\mathcal{L}}(t-1) + 2 \sqrt{\sigma^2_{\mathcal{L}}(t-1)}$. Upon detection of a boundary, we suppress the check until $1000$ samples have been seen after the last detected shift (creating a refractory period to allow the competitive learning model, described next, to acquire enough data to learn).

%

\paragraph{Task Recognition through Competitive Learning: }
To conduct task recognition, $\mathcal{T}(\mathbf{x})$ first randomly projects the input $\mathbf{x}$ to a low-dimensional space (``key'') $\mathbf{k} = \mathbf{R} \cdot \mathbf{x}$ ($\mathbf{R}$ is initialized from a Gaussian distribution). 
We then update a rolling average estimate of the streaming input using this generated key as follows: $\mathbf{k}_\mu = (1 - \tau) \mathbf{k}_\mu + \tau \mathbf{k}$ (with $\tau = 0.65$). Finally, with the matrix $\mathbf{Q}$, FNBG maps this rolling representation $\mathbf{k}_\mu$ to a decision as to which task context the S-NCN generative cortex is to utilize, i.e., $\mathbf{\hat{t}} = \Call{bkWTA}{\mathbf{Q} \cdot (\mathbf{k}_\mu/||\mathbf{k}_\mu)||_2, K=1}$.
The task pointer produced for the S-NCN model is then $t_i = \arg\max(\mathbf{\hat{t}})$ ($\arg\max(\mathbf{\hat{t}})$ returns the index of the maximum value in $\mathbf{\hat{t}}$).

To update the FNBG weights, while also avoiding catastrophic interference in $\mathcal{T}(\mathbf{x})$ itself, we propose a biologically-inspired learning rule based on competitive learning. Specifically, we develop what we call ``guided competitive learning'', since during the act of processing a stream of certain samples from the task we know we are operating on, we know which neuron out of a set of $T-1$ task output neurons should be selected. This leads to the following update rule
formally defined as:
\begin{align}
    \Delta \mathbf{Q} = -(\rho \mathbf{t}) \cdot (\mathbf{k}_\mu/||\mathbf{k}_\mu||_2)^T \label{eqn:fnbg_update}
\end{align}
where $\rho$ is the competitive weight adjustment factor (a value we found works well in the range of $[0.5, 1)$). Note that $\mathbf{t} = \mathbf{\hat{t}}$ recovers an unsupervised classical competitive Hebbian learning. However, we force the model to a specific task pointer value by using $\mathbf{t}$, the one-hot encoding of the dynamic integer variable $t$ maintained by the FNBG itself, initialized as $t = 0$.
Every time a task shift is detected according to Equations \ref{eqn:shift_delta}-\ref{eqn:shift_var}, this dynamic variable is incremented by one, i.e., $t \leftarrow t + 1$.

For both task recognition and the FNBG's synaptic update, note that the rolling representation $\mathbf{k}_\mu$ is normalized by its Euclidean norm so that we may utilize a vectorizable form of competitive learning based on dot products (taking advantage of GPU-based matrix multiplication). In short, we take the (normalized) rolling average representation of the input stream for a given task, compute its dot-product simultaneously with all currently-available task weight vectors, and choose the dot product with maximal value as the winner. Finally, 
we re-normalize the matrix $Q$ by its column Euclidean norms after each update, i.e., $Q[:,i] = (Q[:,i] + \Delta Q[:,i])/||Q[:,i]||_2$ where $Q[:,i]$ indicates extracting all values in column $i$ from $Q$ (this normalization is similar to that of adaptive resonance theory \cite{grossberg2013adaptive}).

\subsection{Putting It All Together: A Complementary System}
At a high level, given the above, the full S-NCN complementary system, depicted in Figure \ref{fig:ncn} (Right), consists of:
1) a task selection model (inspired by the information routing/executive control behavior of the basal ganglia \cite{leisman2014cognitive}) which creates the task contexts that laterally inhibit/gate the activities of the generative S-NCN, and
2) a generative model that learns to predict inputs/labels given a task context. 
In essence, the FNBG $\mathcal{T}(\mathbf{x})$ takes in $\mathbf{x}_j$ to produce $\mathbf{\hat{t}}$ (a one-hot encoding of the task pointer $t_i$) which is then fed into the generative S-NCN (along with $\mathbf{x}$ and $\mathbf{y}$) to compute predictions. 
\footnote{Please see the Appendix where we summarize symbols, notation, and abbreviation used in this work.}

\section{Experiments}
\label{sec:exp}

\paragraph{Simulation Setup: }
\label{sec:setup}
In our experiments, we train models with three hidden layers, whether they be multilayer perceptrons (MLPs) or S-NCNs and compare against baselines from the literature. All models were restricted to contain \textcolor{black}{(a maximum of)} $500$ units per layer. \textcolor{black}{For the S-NCN,} weights were initialized from a Gaussian distribution scaled by each layer's fan-in \textcolor{black}{and were optimized using stochastic gradient descent with learning rate of $\lambda = 0.01$}.
Baseline models were trained on each task for $40$ epochs while the S-NCN only made a single pass. 
The output layer for each MLP was a maximum entropy classifier and the objective was to minimize Categorical cross entropy -- in the S-NCN, this was encoded in its label error neurons $\mathbf{e}^0_y$. (See Appendix for experimental details, computing infrastructure, hyper-parameter \textcolor{black}{details}, and code details.)

\begin{table*}[!t]
\begin{center}
\caption{Generalization metrics (10 trials) for sequence orderings \# 1 \& \#2 (higher values are better).}
\label{results:seq_variant}
\footnotesize
\begin{tabular}{|l||c|c||c|c|}
\hline
  \multicolumn{1}{|l||}{}&\multicolumn{4}{c|}{\begin{tabular}[x]{@{}c@{}}\textbf{Ordering \#1: $\{M1, M2, GD1, FM1, FM2, GD2\}$ (High Color Sim.)}\\\end{tabular}} \tabularnewline
  \multicolumn{1}{|l||}{}&\multicolumn{2}{c}{\begin{tabular}[x]{@{}c@{}}\textbf{Equal}\\\end{tabular}}&\multicolumn{2}{c|}{\begin{tabular}[x]{@{}c@{}}\textbf{Unequal}\\\end{tabular}}\tabularnewline
  & \textbf{ACC} & \textbf{BWT} & \textbf{ACC} & \textbf{BWT} \\
  \hline
  Backprop & $0.241\pm0.050$ & $-0.759\pm0.030$ & $0.185\pm0.048$ & $-0.791\pm0.048$ \\
  EWC & $0.280\pm0.023$ & $-0.714\pm0.030$ & $0.185\pm0.046$ & $-0.726\pm0.039$ \\
  Md-IMM & $0.521\pm0.027$ & $-0.392\pm0.023$ & $0.480\pm0.039$ & $-0.240\pm0.040$ \\
  DT+Md-IMM & $0.530\pm0.024$ & $-0.387\pm0.021$ & $0.551\pm0.042$ & $-0.220\pm0.042$ \\
  L2+DT+Md-IMM & $0.532\pm0.025$ & $-0.237\pm0.027$ & $0.520\pm0.040$ & $-0.240\pm0.045$ \\
  HAT & $0.550\pm0.019$ & $-0.211\pm0.020$ & $0.492\pm0.031$ & $-0.231\pm0.036$ \\
  \hline
  S-NCN (ours) & $\mathbf{0.716 \pm 0.013}$ & $\mathbf{-0.031 \pm 0.017}$ & $\mathbf{0.713 \pm 0.011}$ & $\mathbf{-0.041 \pm 0.012}$ \\
  \hline
  \hline 
  \multicolumn{1}{|l||}{}&\multicolumn{4}{c|}{\begin{tabular}[x]{@{}c@{}}\textbf{Ordering \#2: $\{GD2, M1, FM2, M2, GD1, FM1\}$ (Low Color Sim.)}\\\end{tabular}} \tabularnewline
  \hline
  Backprop & $0.303\pm0.030$ & $-0.644\pm0.037$ & $0.287\pm0.043$ & $-0.671\pm0.043$ \\
  EWC & $0.303\pm0.031$ & $-0.643\pm0.033$ & $0.291\pm0.039$ & $-0.663\pm0.047$ \\
  Md-IMM & $0.584\pm0.027$ & $-0.091\pm0.030$ & $0.533\pm0.034$ & $-0.230\pm0.036$ \\
  DT+Md-IMM & $0.591\pm0.020$ & $-0.088\pm0.032$ & $0.528\pm0.036$ & $-0.211\pm0.039$ \\
  L2+DT+Md-IMM & $0.630\pm0.029$ & $-0.076\pm0.030$ & $0.551\pm0.037$ & $-0.201\pm0.041$ \\
  HAT & $0.596\pm0.026$ & $-0.114\pm0.029$ & $0.563\pm0.031$ & $-0.210\pm0.044$ \\
  \hline
  S-NCN (ours) & $\mathbf{0.721 \pm 0.014}$ & $\mathbf{-0.042 \pm 0.013}$ & $\mathbf{0.667 \pm 0.011}$ & $\mathbf{-0.097 \pm 0.013}$ \\
  \hline
\end{tabular}
\end{center}
\vspace{-0.5cm}
\end{table*}

\paragraph{Evaluation Metrics:} %
\label{sec:metrics}
To measure model generalization over the sequence of tasks, we make use of the resulting task matrix $R$ (as in \cite{lopez2017gradient}), an $N \times N$ matrix of task accuracy scores (normalized to $[0,1]$), where in this study $N = T$. We measure average accuracy (ACC) (mean performance across tasks) and backward transfer (BWT). BWT measures the influence that learning a task $T_t$ has on the performance of task $T_k < T_t$. A positive BWT indicates that a learning task $T_t$ increases performance on a preceding task $T_k$. As such, a higher BWT is better and a strongly negative BWT means there is stronger (more catastrophic) forgetting. Mean and standard deviation ($10$ trials) are reported for ACC and mean ($10$ trials) for BWT (see Appendix for standard deviation).
The formulas for ACC, BWT, and for a new set of metrics created to analyze memory retention, are provided in the Appendix. 

\begin{table*}[!t]
\centering
\caption{Generalization metrics (10 trials) for Split MNIST, Split Fashion MNIST (FMNIST) and Not-MNIST benchmarks. \textcolor{black}{Above dashed line are multi-head models \& below are single-head models.}
}
\label{results:benchmarks}
\footnotesize
\begin{tabular}{|l||c|c||c|c|c|c|}
\hline
  \multicolumn{1}{|l||}{}&\multicolumn{2}{c}{\begin{tabular}[x]{@{}c@{}}\textbf{MNIST}\\\end{tabular}}&\multicolumn{2}{c|}{\begin{tabular}[x]{@{}c@{}}\textbf{Fashion MNIST}\\\end{tabular}}&\multicolumn{2}{c|}{\begin{tabular}[x]{@{}c@{}}\textbf{NotMNIST}\\\end{tabular}}\tabularnewline
  & \textbf{ACC} & \textbf{BWT} & \textbf{ACC} & \textbf{BWT} & \textbf{ACC} & \textbf{BWT}\\
  \hline
  \hline
  EWC & $0.760 \pm 0.030$ & $-0.210$ & $0.739 \pm 0.020$ & $-0.201$ & $0.790 \pm 0.020$ & $-0.176$ \\
  VCL & $0.980 \pm 0.210$ & $-0.003$  & $0.980 \pm 0.20$ & $-0.002$ & $0.953 \pm 0.003$ & $-0.004$\\
  IMM & $0.951 \pm 0.018$ & $-0.007$ & $0.950 \pm 0.013$ & $-0.005$ & $0.925 \pm 0.011$ & $-0.006$ \\
  HAT & $0.972 \pm 0.011$ & $-0.040$  & $0.968 \pm 0.011$ & $-0.004$ & $0.942 \pm 0.009$ & $-0.005$  \\
  GEM & $0.922 \pm 0.110$ & $\mathbf{+0.001}$  & $0.930 \pm 0.12$ & $+0.001$ & $0.919 \pm 0.021$ & $-0.003$  \\
  \textcolor{black}{A-GEM} & \textcolor{black}{$0.950 \pm 0.09$} & \textcolor{black}{$\mathbf{+0.001}$}  & \textcolor{black}{$0.955 \pm 0.11$} & \textcolor{black}{$+0.001$} & \textcolor{black}{$0.925 \pm 0.020$} & \textcolor{black}{$-0.002$}  \\
  \textcolor{black}{ER} & \textcolor{black}{$0.938 \pm 0.06$} & \textcolor{black}{$\mathbf{-0.002}$}  & \textcolor{black}{$0.945 \pm 0.10$} & \textcolor{black}{$-0.004$} & \textcolor{black}{$0.927 \pm 0.016$} & \textcolor{black}{$-0.004$}  \\
  \hdashline 
  EWC & $0.190 \pm 0.030$ & $-0.357$ & $0.199 \pm 0.06$ & $-0.350$ & $0.186 \pm 0.020$ & $-0.361$  \\
  NR+M & $0.950 \pm 0.470$ & $-0.100$  & $0.948 \pm 0.380$ & $-0.090$ & $0.880 \pm 0.028$ & $-0.103$   \\
  SI & $0.197 \pm 0.110$ & $-0.367$  & $0.198 \pm 0.100$ & $-0.370$ & $0.161 \pm 0.030$ & $-0.370$  \\
  MAS & $0.195 \pm 0.290$ & $-0.340$  & $0.180 \pm 0.250$  & $-0.340$ & $0.178 \pm 0.060$ & $-0.341$  \\
  Lwf & $0.846 \pm 0.340$ & $-0.120$  & $0.875 \pm 0.300$ & $-0.130$ & $0.626 \pm 0.091$ & $-0.130$ \\
  ICarl & $0.940 \pm 0.410$ & $-0.100$  & $0.960 \pm 0.400$ & $-0.110$ & $0.887 \pm 0.102$ & $-0.109$  \\
  Lucir & $0.940 \pm 0.310$ & $-0.103$   & $0.950 \pm 0.340$ & $-0.110$ & $0.935 \pm 0.093$ & $-0.101$  \\
  \textcolor{black}{GDumb} & \textcolor{black}{$0.978 \pm 0.09$} & \textcolor{black}{$\mathbf{-0.005}$}  & \textcolor{black}{$0.973 \pm 0.09$} & \textcolor{black}{$-0.006$} & \textcolor{black}{$0.940 \pm 0.080$} & \textcolor{black}{$-0.004$}  \\
  Mnem  & $0.960 \pm 0.320$ & $-0.091$  & $0.968 \pm 0.300$ & $\mathbf{+0.007}$ & $0.950 \pm 0.071$ & $-0.011$  \\
  \hline
  S-NCN & $\mathbf{0.981 \pm 0.300}$ & $-0.005$   & $\mathbf{0.982 \pm 0.400}$ & $-0.003$ & $\mathbf{0.957 \pm 0.400}$ & $\mathbf{-0.004}$ \\
  \hline
\end{tabular}
\vspace{-0.45cm}
\end{table*}

\subsection{The Multi-Dataset Task Stream}
\label{sec:dataset_creation}
To start, we tested the S-NCN model on a complex task sequence composed of several learning benchmarks.
We create task sequences by breaking apart MNIST (\emph{M}), Fashion MNIST (\emph{FM}), and Google Draw (\emph{GD}) each into two ``sub-tasks'' (e.g., for MNIST, \emph{M1} and \emph{M2}), or portions of data with a particular subset of the original dataset's classes. See the Appendix for details on sub-tasks/task sequence creation. In Table \ref{results:seq_variant}, we present two task orderings (Ordering \#1 is ``High Color Sim.'' and Ordering \#2 is ``Low Color Sim.'') each under two conditions: sub-tasks that have 1) an equal number of classes ($5$ each), and 2) an unequal number of classes (number of classes was chosen randomly, omitting $5$ as an option). The number of classes was sampled once and held constant for all trials.

We evaluate our proposed S-NCN system (as well as four variations of it in the Appendix) -- hyperparameters were $\beta = 0.05$, $K = 10$, $\eta_g = 0.9$, $\eta_e = 0.01$, $\alpha = 0.98$).
The baselines include an MLP trained only with backprop (Backprop), Elastic Weight Consolidation (EWC) \cite{kirkpatrick2017overcoming}, the Mode-IMM method \cite{forgetting_iim}, the Md-IMM method combined with either DropTransfer (DT+Md-IMM) or both L2-Transfer and DropTransfer (L2-DT-Mode-IMM) \cite{forgetting_iim}, and the competitive model, hard attention to task (HAT) \cite{forgetting_hard_attention}.
For each baseline, we tuned hyper-parameters based on their accuracy on each task's development set. See Appendix for extra baseline results.

\paragraph{Discussion: }
Results are reported in Table \ref{results:seq_variant} 
(see Appendix for more results). Each simulation was run $10$ times (each trial used a unique seed) -- we report both mean and standard deviation.
As observed in our results, \textbf{we see that our S-NCN model  outperforms all baselines consistently, in terms of ACC and BWT, exhibiting improved memory retention over the baselines, such as backprop, and more notably, EWC} (and, in the Appendix, our expanded results show that the FNBG-driven lateral inhibition is key to improving memory retention the most).
This result is robust across both task sequences and equal/unequal class settings. Meta-parameters for the S-NCNs were only tweaked minorly, with the same values used in all scenarios. The observation that lateral inhibition improves the neural computation (in our model) also corroborates the result of \cite{oreilly2001inhibition}.

\subsection{Continual Learning Benchmarks}
\label{sec:benchmarks}
To connect our model to current lifelong learning results, we experimented with a wide swatch of approaches on three benchmarks -- Split MNIST, Split NotMNIST, and Split Fashion MNIST (FMNIST). Furthermore, we compare to multi-head (below dashed line in Table \ref{results:benchmarks}) and single-head models (above dashed line). 
We compare the S-NCN 
to replay/rehearsal and non-replay methods:
na\"{\i}ve rehearsal with memory (NR+M), EWC, 
synaptic intelligence (SI) \cite{zenke2017continual}, MAS \cite{mas}, Lwf \cite{lwf}, GEM \cite{gem}, ICarl \cite{icarl}, Lucir \cite{lucir}, and Mnemonics \cite{mnemonics} (additional baselines can be found in the Appendix). 

\paragraph{Discussion: } In Table \ref{results:benchmarks}, we report ACC and BWT, averaged over $10$ trials, offering an extensive, comprehensive comparison of methods and demonstrating that, for all three benchmarks, \textbf{the proposed S-NCN outperforms all of them in terms of ACC}, and nearly all in terms of BWT (and on par with GEM, the difference in BWT being negligible) demonstrating the power afforded by designing models with stronger grounding in neurobiology (see Appendix for a discussion of limitations). Furthermore, it is promising to see that the S-NCN outperforms/matches performance with not only the single-head models but also with the multi-head models (except GEM), which enjoy an easier version of the problem since they can utilize a different classifier per task. Finally, note that the S-NCN, due to the FNBG-driven lateral competition, learns to compose task contexts in a data-dependent manner.
Desirably, the S-NCN is a single-head model, meaning that it does not grow out a separate softmax classifier for each task (as in multi-headed approaches, e.g., HAT, IMM, GEM), which means that the S-NCN is tackling the harder form of the forgetting problem, learning representations that preserve knowledge across disjoint tasks. Furthermore, note that our model is online, whereas models such as IMM or SI require multiple passes per task dataset, and does not require validation data in order to run an expensive neural architecture search (NAS) as in \cite{li2019learn}.

\section{Conclusion}
\label{sec:conclusion}
In this paper, we proposed the sequential neural coding network (S-NCN), an interactive generative model, and its local learning procedure for lifelong machine learning. As demonstrated on several benchmarks/setups, this model retains the knowledge acquired from prior tasks when learning new ones in  task data streams, primarily when lateral inhibition, driven by a self-organizing task selection model, sharpens neural activities within its layers.
In terms of negative social impact, S-NCN-based lifelong learning models could lead to the development of better-performing systems, i.e., robotic systems (drones), that might result in the loss of life (see Appendix for an expanded discussion). To safeguard against this, it will be important to develop an ethical framework to guide the design and training of general intelligent systems to ensure safe integration into human society.

\bibliographystyle{IEEEtran}
\bibliography{ref}


\section{Appendix}
\input{appendix_arxiv}


\end{document}

%% file: appendix_arxiv.tex
\section*{Computational Resource Setup} 
\label{sec:hardware_setup}

All experiments were performed with 128GB RAM on an intel Xeon server with 3.5GHZ processors, consisting of 4 1080Ti GPUs. Any of our models  can easily fit into single 1080Ti GPUs -- a multi GPU setup was only used to speed up the computation. All models are coded in Tensorflow 2.2 (with eager mode) and we only use basic parallelism provided by the Tensorflow library/package to speed up the computation.

\noindent
\textbf{Experimental Code:} Code will be released to reproduce the results of this paper upon acceptance of publication (the GitHub link will be added to this appendix). Support for applying the S-NCN system to other datasets will be provided to facilitate further research from the machine learning community.

\section*{On Negative Societal Impacts}

The potential negative societal impact of the proposed lifelong learning system is indirect -- while the model and algorithmic framework we develop is foundational in nature, it could potentially affect the myriad of applications/systems currently in development today or to be developed in the future. As a result, at best, the same negative consequences that result from using and integrating backprop-based ANNs are still present when using S-NCN training process instead. At worst, given that we have presented promising results on several lifelong learning benchmarks, the S-NCN could facilitate the development of potentially better-performing robotic agents that might be used in military applications that might result in the loss of life, such as drones. Despite the benefits that S-NCN offers to the statistical learning and cognitive neuroscience communities, one should consider the drawbacks of building powerful neural systems to drive applications.

\section*{Remarks: On the Details of Sequential Neural Coding}

\textbf{Definition Table:} In Table \ref{table:definitions}, we explain what each mathematical symbol/operation/abbreviation in the main paper represents.

\textcolor{black}{
\textbf{Derivation of State \& Weight Updates:} As mentioned in the main paper, the S-NCN's generative circuit minimizes an objective function known as total discrepancy (ToD) when it is presented with input stimuli $(\mathbf{x}_i, \mathbf{y}_i))$. The ToD is formally:
\begin{align}
    \mathcal{L}(\Theta) = \sum^{L-1}_{\ell=0} \frac{1}{2} (|| \mathbf{z}^\ell - \mathbf{z}^\ell_{\mu} ||_2)^2
    = \sum^{L-1}_{\ell=0} \frac{1}{2} \sum_j \big( \mathbf{z}^\ell[j] - \mathbf{z}^\ell_{\mu}[j] \big)^2
    \label{eqn:tod}
\end{align}
where $\mathbf{z}^\ell[j]$ means that we extract the $j$-th element of vector $\mathbf{z}^\ell$ (and we have simplified the expression by squaring the square root operator of the L2 norm, giving us a sum of squared dimensions).
Since all of the latent states of the generative circuit are continuous, the updates will follow the form of the exact gradient, i.e., differentiation (which would permit the use of gradient descent), to optimize the latent variables and the synaptic weight parameters. Given this, the partial derivative of Equation \ref{eqn:tod} with respect to any layer of neural activities (or latent state) $\mathbf{z}^\ell$ would be:
\begin{align}
     \frac{\partial \mathcal{L}(\Theta)}{\partial \mathbf{z}^\ell} &=  
                    \left( \frac{\partial \mathbf{z}^{\ell-1}_\mu}
                    {\partial \mathbf{z}^\ell} \cdot \Big( 
                    (\mathbf{z}^{\ell-1}-\mathbf{z}^{\ell-1}_\mu) \Big) \right)-
                    (\mathbf{z}^{\ell}-\mathbf{z}^{\ell}_\mu) \label{eqn:latent_update1}\\
    &= \Big[ (\mathbf{W}^\ell)^T \cdot  
                    (\mathbf{z}^{\ell-1}-\mathbf{z}^{\ell-1}_\mu) \Big] \odot \frac{\partial \phi^\ell(\mathbf{z}^\ell)}{\partial \mathbf{z}^\ell}  - (\mathbf{z}^{\ell}-\mathbf{z}^{\ell}_\mu) \label{eqn:latent_update2} \\
    &= (\mathbf{W}^\ell)^T \cdot (\mathbf{e}^{\ell-1}) \odot \frac{\partial \phi^\ell(\mathbf{z}^\ell)}{\partial \mathbf{z}^\ell}  - \mathbf{e}^\ell \label{eqn:latent_update3}
\end{align}
where we notice that the error neurons are derived directly from the ToD objective as well, i.e., $\mathbf{e}^\ell = \frac{\partial \partial \mathcal{L}(\Theta)}{\partial \mathbf{z}^\ell_\mu} = \mathbf{z}^\ell - \mathbf{z}^\ell_\mu$ (allowing us to write Equation \ref{eqn:latent_update2} in terms of error neurons as in Equation \ref{eqn:latent_update3}). 
Alternatively, by replacing the term $\frac{\partial \mathbf{z}^{\ell-1}}{\partial \mathbf{z}^\ell}$ with a learnable error matrix $\mathbf{E}^\ell$ instead, Equation \ref{eqn:latent_update2} can be simplified to the following:
\begin{align}
    \frac{\partial \mathcal{L}(\Theta)}{\partial \mathbf{z}^\ell} \approx \mathbf{d}^\ell =  \mathbf{E}^\ell \cdot \mathbf{e}^{\ell-1} - \mathbf{e}^\ell \label{eqn:z_l_delta}
\end{align}
which is a stable derivative-free perturbation $\mathbf{d}^\ell$ (so long as the activation function $\phi^\ell()$ is monotonically increasing)  to the latent neural activities (and as noted in \cite{ororbia2022neural}, the dampening effect that the activation function derivative $\frac{\partial \phi^\ell(\mathbf{z}^\ell)}{\partial \mathbf{z}^\ell}$ would have can be approximated with biologically-plausible dampening functions if needed). The final update to the latent neural activities can then be performed using a gradient-ascent like operation, i.e.,  $\mathbf{z}^\ell(k) = f^\ell( \mathbf{z}^\ell(k-1) + \beta \mathbf{d}^\ell )$, which is what was presented in the main paper.
}

\begin{table}[!t]
\caption{Table of key symbol/operator/abbreviation definitions.}
\label{table:definitions}
\begin{center}
\begin{tabular}{||c c||} 
 \hline
 \textbf{Item} & \textbf{Explanation} \\ [0.5ex] 
 \hline\hline
 S-NCN & Sequential neural coding network (model) \\ 
 \hline
 FNBG & Functional neural basal ganglia (model) \\
 \hline
 $\mathbf{v} \in \mathcal{R}^{D \times 1}$ & A column vector $\mathbf{v}$ of shape $D \times 1$ \\
 \hline
 $\mathbf{M} \in \mathcal{R}^{B \times D}$ & A matrix $\mathbf{M}$ of shape $B \times D$ \\
 \hline
 $\cdot$ & Matrix/vector multiplication \\
 \hline
 $\odot$ & Hadamard product (element-wise multiplication) \\
 \hline
 $(\mathbf{v})^T$ & Transpose of $\mathbf{v}$ \\
 \hline
 $||\mathbf{v}||_2$ & Euclidean norm of $\mathbf{v}$ \\
 \hline
 $\mathbf{x}_j$ & The $j$th data point (image) sampled from task $\mathcal{T}_i$ \\
 \hline
 $\mathbf{y}_j$ & The $j$th data label (one-hot encoded) sampled from task $\mathcal{T}_i$ \\
 \hline
 $P_\ell$ & The $\ell$-th predictor/layer of the S-NCN generative circuit. \\
 \hline
 $J_x$ & Dimensionality of input $\mathbf{x}_j$ \\
 \hline
 $J_\ell$ & Number of neurons in layer $P_\ell$ of the S-NCN generative circuit. \\
 \hline
\end{tabular}
\end{center}
\vspace{-0.5cm}
\end{table}

\textcolor{black}{
Deriving the updates to the synaptic generative parameters is also done in a similar fashion as above, i.e., by taking the gradient of ToD with respect to $\mathbf{W}^\ell$.
\begin{align}
    \frac{\partial \mathcal{L}(\Theta)}{\partial \mathbf{W}^\ell} \propto \Delta \mathbf{W}^\ell &=\frac{\partial \mathcal{L}(\Theta)}{\partial \mathbf{z}^\ell_\mu} \cdot  \left(\phi^{\ell+1}(\mathbf{z}^{\ell+1})\right)^T, \mbox{where, } \mathbf{z}^\ell_\mu = \mathbf{W}^\ell \cdot \phi^{\ell+1}(\mathbf{z}^{\ell+1})  \\ 
    &= (\mathbf{z}^\ell - \mathbf{z}^\ell_\mu) \cdot (\phi^{\ell + 1}( \mathbf{z}^{\ell+1}) )^T = \mathbf{e}^\ell \cdot (\phi^{\ell + 1}( \mathbf{z}^{\ell+1}) )^T \mbox{.} \label{eqn:W_l_update}
\end{align}
If we are using $\mathbf{E}^\ell$ feedback/error matrices (as we do in this paper), we can leverage a simple Hebbian update $\Delta \mathbf{E}^\ell = \alpha \Big( \phi^{\ell+1}(\mathbf{z}^{\ell+1}) \cdot  (\mathbf{e}^\ell)^T \Big)$ \cite{ororbia2022neural} (which, if applied to $\mathbf{E}^\ell$ every time that Equation \ref{eqn:W_l_update} is applied to $\mathbf{W}^\ell$, allows $\mathbf{E}^\ell$ to converge to the approximate transpose of $\mathbf{W}^\ell$). Much as was done for the states, synaptic weight matrices are updated via gradient ascent:  $\mathbf{W}^\ell = \mathbf{W}^\ell + \lambda \Delta \mathbf{W}^\ell$ and $\mathbf{E}^\ell = \mathbf{E}^\ell + \lambda \Delta \mathbf{E}^\ell$ ($\lambda$ is the learning rate/step size).
}

\textcolor{black}{
\textbf{How would a model with symmetric connections behave?} 
A model without separate feedback connections (in contrast to the S-NCN we experiment that uses asymmetric forward/feedback weights) would behave quite similarly yet favorably offer a reduction in memory cost (one does not need to store separate feedback/error matrices in memory). In other words, one could certainly swap out $\mathbf{E}^\ell$ with $(\mathbf{W}^\ell)^T$ if this memory cost reduction was desired/necessary. However, by utilizing separate learnable feedback synapses, the S-NCN in the form presented in this study resolves the weight transport problem, a well-known biological criticism of backprop where error/teaching information is carried backwards along the same synapses that were used to forward propagate information.
Interestingly enough, in preliminary experimentation, we found that using separate learnable feedback synapses improved the generative modeling/reconstruction ability of the generative cortex (particularly in the online case). Although we will investigate this effect in future work, we note this change in generative performance did not really seem to impact the classification accuracy (arguably because discrimination is easier than generation). }

\textcolor{black}{
\textbf{Initializing Latent States:} In the S-NCN's generative circuit, there are several layers of neural activities that are not clamped to data, e.g., $\mathbf{z}^1,\mathbf{z}^2,...,\mathbf{z}^L$. These activity vectors, as mentioned in the main paper, are initialized to zero vectors (i.e., $\mathbf{z}^\ell = \mathbf{0}$) before they are updated/modified by the message passing that occurs over $K$ steps of processing input stimuli ($\mathbf{x}, \mathbf{y}$). While these initially zero vectors will eventually become non-zero vectors, particularly after a minimum of $K = L$ steps (for example: after $k = 1$, $\mathbf{z}^1$ will be non-zero given that the error neurons in layer $0$ will be non-zero and thus the layer $1$ state perturbation vector $\mathbf{d}^1$ will contain non-zero entries; after $k = 2$, $\mathbf{z}^2$ will contain non-zero values, and so on and so forth), it is entirely possible to randomly initialize these states with non-zero vectors (though it is recommended to keep the magnitude of the randomly chosen initial numbers relatively small). We leave investigation of alternative initialization schemes for future work.
}

\textcolor{black}{
\textbf{Relationship to Surrogate Gradients:} A particular line work that shares interesting relationships with the generative circuit of the S-NCN is that of surrogate gradients, such as decoupled neural interfaces (DNIs)  \cite{jaderberg2017decoupled,czarnecki2017understanding}. In essence, this class of methods aims to resolve one of backprop's central issues -- the update-locking problem (where updates to one layer's synaptic parameters must for other layer's updates to be computed as error/teaching information is backwards propagated down a serial feedback pathway). The key module driving this class of methods is the introduction of a gradient predictor, which can be adapted/taught to approximate actual gradients as produced by backprop, ultimately, after training the predictor's well enough removing the need for backprop later in training and permitting parallel, asynchronous updates to be made to deep, even recurrent, network architectures. 
In contrast to surrogate gradient-based approaches, the S-NCN works to compute synaptic updates without resorting to predicting gradients given that its generative circuit is naturally layer-wise parallelizable. In effect, each layer-wise prediction is made independently of the others (unlike the typical forward passes in modern-day deep networks) and the synaptic updates for each layer (both forward and error/feedback synapses) can be made without others having been computed/completed. This opens the door to potential parallel asynchronous implementations of the S-NCN that could drastically speed up its computation further. In contrast to DNIs, the S-NCN’s generative cortex does not require training gradient predictors (DNIs typically require access to true gradients provided by backprop in order to train them properly) and, without incurring synthetic approximation errors (as in a fully-unlocked network using DNI, where now even the layer activities require additional modules to be trained to predict actual layer-wise activities) furthermore, resolves both update and forward locking problems. Crucially, the S-NCN’s updates are biologically plausible – it only requires simple (multi-term) Hebbian updates for the generative cortex and competitive Hebbian updates for the basal ganglia.
}

\textbf{General S-NCN System Process Intuition:} From a high-level intuitive point-of-view, \textcolor{black}{the S-NCN system described in the main paper is composed of two complementary neural circuits: 1) an interactive generative (cortical) circuit that learns to predict its input stimuli (pixel images and their respective labels), and 2) the functional neural basal ganglia (FNBG) which is a specialized circuit that learns to group pixel images into unique ``task'' contexts. }
When presented with a sample or mini-batch at any point within the task stream being sampled, the S-NCN does the following: 
\begin{enumerate}[noitemsep,nolistsep]
    \item The FNBG task selection model determines if the currently sampled data is coming from the same task that the S-NCN has currently been processing or if it comes from either new/different task. (Note, not mentioned in the paper, the FNBG can also determine that the current data belongs to a task that it has previously seen by letting its set of neurons compete and determining if the winner has a very high dot product score -- usually checked against threshold). Before letting the S-NCN generative circuit process and learn from the current data, if the FNBG determines that the data is coming from a novel task, it will create a new task context memory $\mathbf{g}^\ell_t$ for each $\mathbf{M}^\ell$ which will subsequently drive the S-NCN as it processes the current data.
    \item Given the task context provided by the FNBG, the S-NCN generative circuit will then process the current data over a fixed stimulus window (or for $K$ discrete time steps), adjusting its synapses so that it may better generate the input sensory sample better in the future as well as predict its label. Note that the update to the S-NCN's synapses will also trigger an update to the task context memory vectors as well as an update to the FNBG's synapses (which are themselves adjusted through competitive Hebbian learning).
\end{enumerate}
\textbf{Task Shift Detection Intuition:} To provide further intuition as to how the FNBG task selector operates when detecting the occurrence of a novel task, i.e., ``task shift detection'', we explain what the key equations presented in the main paper are doing: 
\begin{enumerate}[noitemsep,nolistsep]
    \item The FNBG lets the S-NCN generative circuit continue as it normally would and have it make a prediction of both $\mathbf{x}_j$ and $\mathbf{y}_j$. Then, it extracts the label error neuron vector $\mathbf{e}^0_y$ from the generative model and computes its squared Euclidean norm $||\mathbf{e}^0_y||^2_2$.
    \item The FNBG maintains two particular scalar parameters $\mu_{\mathcal{L}}$ and $\sigma^2_{\mathcal{L}}$, which are the running mean and variance of the squared Euclidean norm of the S-NCN's label error neuron activity level. The key Equations 7-9 in the main paper depict how the FNBG updates these mean $\mu_{\mathcal{L}}(t)$ and variance $\sigma^2_{\mathcal{L}}(t)$ parameters (based on Welford's algorithm for calculating an online mean and variance).
    \item Once the FNBG updates these two statistical parameters, it then performs a check of its current new mean $\mu_{\mathcal{L}}(t)$ against the previous value ($t-1$) of its mean and variance parameters, i.e., $\mu_{\mathcal{L}}(t-1)$ and $\sigma^2_{\mathcal{L}}(t-1)$. This check is specifically, as presented in the main paper, as follows: $\mu_{\mathcal{L}}(t) >  \mu_{\mathcal{L}}(t-1) + 2 \sqrt{\sigma^2_{\mathcal{L}}(t-1)}$ which says that, if the current mean of the label error neuron activity (at time $t$) is greater than the sum of the previous mean (at time $t-1$) plus two times the previous standard deviation, the S-NCN system has encountered a \emph{shift in task}, which means that, if this inequality evaluates to true, the currently encountered data/mini-batch comes from either a novel task or a previously encountered task (but not from the current task). If this task shit inequality evaluates to true, the FNBG will not allow the S-NCN generative circuit to update its synapses but instead force it to recompute its prediction of the current data but using a newly generated task context. Otherwise, if the inequality evaluates to false, then the FNBG will let the S-NCN generative circuit continue and update its synaptic parameters.
\end{enumerate}

\textbf{Design Motivation/Intuition:} The key motivations behind the S-NCN's design are:
1) to develop a neurobiologically-inspired online approach to learn a generative model of $p(\mathbf{x},\mathbf{y})$ (one motivated by perceptual cortical circuits), 
2) to develop an online information routing model (one motivated/inspired by the basal ganglia) that suppresses/drives the neural activities in the generative model of (1) depending on the task that it decides that the system is operating on -- this crucially removes the need for user-provided task descriptors. In addition, by design, the S-NCN: 1) exhibits no need for activation function derivatives, 2) exhibits no update-locking (i.e., it is layer-wise parallelizable), 3) does not require/need a global feedback pathway to drive/facilitate credit assignment (i.e., side-stepping exploding/vanishing gradient problems), and 4) it resolves the weight transport problem by using asymmetric generative and error-correction synapses (those these could be tied/shared to reduce the S-NCN's memory footprint if need be). 
In contrast to many current lifelong learning approaches, the S-NCN differs in that it is a complementary system that focuses on the relationship between the basal ganglia (as an information router) and (generative/predictive) cortical regions. This is what allows the S-NCN system to offer the advantage of internal, automatic task selection and task boundary detection, which few modern methods provide.

Although the S-NCN shares predictive processing's (PP's) iterative inference/learning (note that \cite{salvatori2021associative} focused on PP's auto-association abilities) and (note that \cite{ororbia2022neural} focused on PP's generative/sampling abilities); its ability to combat forgetting comes from the interaction between the task selector and the generative circuit (the former drives lateral suppression/excitation in the latter). Note that the S-NCN's generative model could be useful for longer task streams via layer-wise generative replay (each layer could refresh itself in a sleep phase).
This we observer is additional, untapped potential for using the S-NCN's learned directed generative model in a spirit similar to the backprop-based model of \cite{van2020brain}. This we argue could prove useful for longer task sequences with far more tasks.

%

Empirically, we note that asymmetric forward and backwards (error) weights were found to work best for NCN systems without activation derivatives in their state updates, yielding fewer inference steps. However, since backwards weights converge to the approximate transpose of the forward ones, using symmetric weights would reduce memory usage. Note that even fixed, random backward synapses (as in feedback alignment) would work as well (even though we do not explore them in this work), which we found, in preliminary experimentation, reduced ACC by only a few points.

\textcolor{black}{
\textbf{Relationship to Temporal Neural Coding:} Prior related on temporal neural coding \cite{ororbia2017learning,ororbia2018continual} has investigated the design and development of predictive coding frameworks for handling the modeling of time-varying data points (such as frames in a video). However, in contrast to the generative circuit in this work, these previous models were either not layer-wise parallelizable \cite{ororbia2017learning} or were created under a specialized recurrent formulation of PC for processing data over multiple epochs \cite{ororbia2018continual} (such as bouncing ball or digit videos). These models would not be naturally resistant to forgetting like the S-NCN is, given that it is a complementary system where one circuit aids the other in learning task-sensitive/dependent sub-networks.
Furthermore, this earlier related work \cite{ororbia2017learning,ororbia2018continual} focused on unsupervised sequence modeling and did not investigate discriminative forms of learning (e.g., classification) where forgetting is, in our experience, far faster and stronger to observe. (Note that in \cite{ororbia2018continual}, although some improved memory retention across the three sequence modeling tasks was observed, forgetting over tasks was still apparent, and any improvement was more of a pleasant side-effect rather than the result of a mechanism specialized for safe-guarding against forgetting.)
}

\section*{On Weight Update Modulation}

\textcolor{black}{
As presented in the main paper, the synaptic weight updates for the generative circuit of the S-NCN applies modulation matrices to improve learning stability over time (specifically invoking a form of synaptic scaling -- note that these modulation matrices are not meant to mitigate the occurrence of forgetting).}
The modulation factors are (locally) computed as a function of the synapses that they are ultimately meant to support:
\begin{align}
    \widehat{\mathbf{m}}^\ell_W &= \Sigma^{J_{\ell+1}}_{j=1} \mathbf{W}^\ell[:,j], \; \mbox{and}, \; \mathbf{m}^\ell_W = \min \bigg( \frac{\gamma_s \widehat{\mathbf{m}}^\ell_W}{\max(\widehat{\mathbf{m}}^\ell_W)}, 1 \bigg) \\
    \mathbf{S}^\ell_W &= (\mathbf{W}^\ell * 0 + 1) \odot \mathbf{m}^\ell_W
\end{align}
where $\mathbf{W}[:,j]$ denotes the extraction of the $j$th column of $W$, $\max(\widehat{\mathbf{m}}^\ell_W)$ returns the maximum scalar value of $\widehat{\mathbf{m}}^\ell_W$, and $\gamma_s = 2$. We note that the first two formulae collapse the forward matrix to a column vector of normalized multiplicative weighting factors and the third formula converts the column vector to a tiled matrix of the same shape as $\mathbf{W}^\ell$. 
The error weight modulation factor is computed in fashion similar to that of the forward weights:
\begin{align}
    \widehat{\mathbf{m}}^\ell_E &= \Sigma^{J_{\ell}}_{j=1} \mathbf{E}^\ell[:,j],  \mbox{and}, \mathbf{m}^\ell_E = \min \bigg( \frac{2 \widehat{\mathbf{m}}^\ell_E}{\max(\widehat{\mathbf{m}}^\ell_E)}, 1 \bigg) \\
    \mathbf{S}^\ell_E &= (\mathbf{E}^\ell * 0 + 1) \odot \mathbf{m}^\ell_E
\end{align}
where we observe that modulation factors are computed across the pre-synaptic dimension/side of either matrix $\mathbf{W}^\ell$ or $\mathbf{E}^\ell$. The multiplicative modulation terms come from the insight in neuroscience that synaptic scaling, driven by competition across synapses, serves as a global (negative) feedback mechanism for regulating the magnitude of synaptic adjustments 
\cite{turrigiano2008self,ibata2008rapid,moulin2020synaptic}.
\textcolor{black}{
From a practical standpoint, we found that using the above modulation/scaling factors meant we did not have to craft a synaptic normalization scheme (such as in sparse coding schemes, where the columns/rows of a synaptic matrix must be normalized such that they of unit length each time a synaptic matrix is updated).}

\textcolor{black}{
We remark that the modulation factors we introduce could instead be adapted such that they are useful for mitigating forgetting instead, as has been done in other continual learning approaches \cite{beaulieu2020learning,madireddy2020neuromodulated,imam2020rapid,tsuda2020modeling}.
This could potentially help to reduce the cost for growing out new task contexts each time a new task is encountered. For example, one could potentially adapt the modulation factor matrices to instead be conditioned on the output of the S-NCN's functional neural basal ganglia (or another type of task selector circuit, such as one that mimics the cognitive control capbilities of the prefrontal cortex).}

\section*{On Partial Pattern Completion}


In the event that incomplete input $\mathbf{x}_i$ is provided to the GNCN, i.e., portions of $\mathbf{x}_i$ are masked out by the variable $\mathbf{m} \in \{0,1\}^{J_0 \times 1}$, as mentioned in the main paper, we may infer the remaining portions of $\mathbf{x}_i$ by using the relevant output error neurons of the GNCN and treating the bottom sensory/input layer $\mathbf{z}^0_x$ as a partial latent state. Specifically, we update the missing portions, i.e., $1 - \mathbf{m}$, of $\mathbf{z}^0$ in the following manner:
\begin{align}
    \mathbf{z}^0_x = \mathbf{x} \odot \mathbf{m} + \Big( \mathbf{z}^0_x - \beta \mathbf{e}^0_x \Big) \odot (1 - \mathbf{m})
\end{align}
where $\mathbf{e}^0_x = \mathbf{z}^0_x - \mathbf{z}^0_{\mu,x}$ (error neuron signals related to $\mathbf{x}_i$).

\section*{Parameter Optimization Setup and Baseline Details}

\noindent
\textbf{\textcolor{black}{S-NCN Optimization:}} For the S-NCN, we used SGD with a learning rate of $\lambda=0.01$ (this rate was only minorly tuned on the validation set of the first task in preliminary investigation) \textcolor{black}{and mini-batches each containing $10$ samples.} 
\textcolor{black}{Based on preliminary experiments, the S-NCN, in general, was found to be robust w.r.t. such hyper-parameters. However, the} final meta-parameter values used, i.e., $\beta = 0.05$, $K = 10$, $\eta_g = 0.9$, $\eta_e = 0.01$, $\alpha = 0.98$, were obtained by conducting a grid search (using the validation sets to find best generalization). This meant that we searched over the ranges: $\beta = [0.01, 0.1]$, $K = [5,30]$, $\eta_g = [0.5,1.0]$, $\eta_e = [0.005,0.25]$, $\alpha = [0.5,1.0]$. 

\noindent
\textbf{Baseline Descriptions:} The baselines include an MLP trained exclusively with backprop (Backprop), an MLP trained by backprop but regularized by drop-out (Backprop+DO), Elastic Weight Consolidation (EWC) \cite{kirkpatrick2017overcoming}, EWC further regularized by drop-out (EWC-DO), the mean incremental moment matching method (IMM) or Mean-IMM \cite{forgetting_iim}, the Mode-IMM method \cite{forgetting_iim}, the Mean-IMM method combined with either DropTransfer (DT+Mean-IMM) or L2-transfer (L2-Mean-IMM) or both (L2+DT+Mean-IMM) \cite{forgetting_iim}, the Md-IMM method combined with either DropTransfer (DT+Md-IMM) or L2-Transfer (L2+Md-IMM) or both (L2-DT-Mode-IMM) \cite{forgetting_iim}, and the state-of-the-art competitive model, hard attention to task (HAT) \cite{forgetting_hard_attention}. \textcolor{black}{Furthermore, as mentioned in the main paper, we examine other methods including those based on replay/rehearsal: na\"{\i}ve rehearsal with memory (NR+M), EWC, 
synaptic intelligence (SI), MAS \cite{mas}, Lwf \cite{lwf}, GEM \cite{gem}, ICarl \cite{icarl}, Lucir \cite{lucir}, and Mnemonics \cite{mnemonics}.
With respect to very modern baselines, we also include, in the main paper, results for the greedy sampler and dumb learner (GDumb) \cite{prabhu2020gdumb,mai2022online}, experience replay (ER) \cite{rolnick2019experience,mai2022online}, and average gradient episodic memory (A-GEM) \cite{hu2020gradient,mai2022online}.}

\noindent
\textbf{Baseline Meta-parameter Tuning:} For all baselines we take/use the hyper-parameters from each model/algorithm's source work as a starting point and tuned \textcolor{black}{each}, using grid search, the batch size, learning rate, number of hidden units in each layer of the target MLP classifier, and optimizer choice.
\textcolor{black}{
We tuned across the following ranges: learning rate range was $[3e-5, 0.1]$, the optimizer choice was tuned across the discrete set [``SGD'', ``momentum'', ``Adam'', ``AdamW''], the hidden layer size range was $[128,512]$, and number of layers range was $[2-5]$. 
}
For each baseline, we tuned hyper-parameters based on their accuracy on each task's development set (as mentioned in the next sub-section). For IMM, we used the same settings proposed in the original paper as a starting point \cite{forgetting_iim}. However, we found that HAT \cite{forgetting_hard_attention} was quite sensitive to the choice of its two key hyper-parameters: 1) the stability parameter $\mathrm{s_{max}}$, and 2) the ``compressibility'' parameter $\mathrm{c}$. After extensive tuning, we used $\mathrm{s_{max}}=450$ and $\mathrm{c}=0.78$. 
For other baseline-specific hyperparameters, e.g., A-GEM has a gamma and soft-constraint meta-parameter, we used the best-practice values reported in the literature (as we found that these values worked well in general, even after some preliminary experimentation).

\begin{table}[t]
\begin{center}
\textcolor{black}{
\begin{tabular}{ c|c|c|c } 
 \hline
 \multicolumn{4}{c}{Model General Hyperparameters} \\
 \hline
  & \textbf{MNIST} & \textbf{FMNIST} & \textbf{NotMNIST} \\
 \textbf{Model} & \textit{Configuration} & \textit{Configuration} & \textit{Configuration} \\ 
 \hline
 EWC & \begin{tabular}{@{}c@{}}lr = 1e-3, SGD \\  NH = 1024, NL = 2 \end{tabular} & \begin{tabular}{@{}c@{}}lr = 1e-4, SGD+M \\  NH = 512, NL = 3 \end{tabular} & \begin{tabular}{@{}c@{}}lr = 1e-3, SGD \\  NH = 1024, NL = 2 \end{tabular} \\ 
 VCL & \begin{tabular}{@{}c@{}}lr = 2e-4, Adam \\  NH = 512, NL = 3 \end{tabular} & \begin{tabular}{@{}c@{}}lr = 2e-4, SGD+M \\  NH = 512, NL = 3 \end{tabular} & \begin{tabular}{@{}c@{}}lr = 2e-4, Adam \\  NH = 512, NL = 3 \end{tabular} \\ 
 IMM & \begin{tabular}{@{}c@{}}lr = 2e-3, SGD+M \\  NH = 512, NL = 3 \end{tabular} & \begin{tabular}{@{}c@{}}lr = 1e-5, SGD \\  NH = 512, NL = 3 \end{tabular} & \begin{tabular}{@{}c@{}}lr = 1e-4, SGD+M \\  NH = 512, NL = 3 \end{tabular} \\ 
 HAT& \begin{tabular}{@{}c@{}}lr = 1e-4, SGD \\  NH = 512, NL = 3 \end{tabular} & \begin{tabular}{@{}c@{}}lr = 2e-4, SGD \\  NH = 512, NL = 3 \end{tabular} & \begin{tabular}{@{}c@{}}lr = 2e-4, SGD+M \\  NH = 512, NL = 3 \end{tabular} \\ 
 A-GEM & \begin{tabular}{@{}c@{}}lr = 1e-4, SGD+M \\  NH = 512, NL = 3 \end{tabular} & \begin{tabular}{@{}c@{}}lr = 2e-5, SGD \\  NH = 512, NL = 3 \end{tabular} & \begin{tabular}{@{}c@{}}lr = 2e-4, SGD+M \\  NH = 512, NL = 3 \end{tabular} \\ 
 ER & \begin{tabular}{@{}c@{}}lr = 2e-4, Adam \\  NH = 512, NL = 3 \end{tabular} & \begin{tabular}{@{}c@{}}lr = 1e-4, AdamW \\  NH = 512, NL = 3 \end{tabular} & \begin{tabular}{@{}c@{}}lr = 1e-4, AdamW \\  NH = 512, NL = 3 \end{tabular} \\ 
 \hline
 \hline
 EWC & \begin{tabular}{@{}c@{}}lr = 1e-3, SGD+M \\  NH = 1024, NL = 2 \end{tabular} & \begin{tabular}{@{}c@{}}lr = 1e-3, SGD+M \\  NH = 1024, NL = 2 \end{tabular} & \begin{tabular}{@{}c@{}}lr = 1e-3, SGD \\  NH = 512, NL = 3 \end{tabular} \\ 
 NR+M & \begin{tabular}{@{}c@{}}lr = 1e-4, SGD+M \\  NH = 512, NL = 3 \end{tabular} & \begin{tabular}{@{}c@{}}lr = 1e-3, SGD+M \\  NH = 512, NL = 3 \end{tabular} & \begin{tabular}{@{}c@{}}lr = 1e-4, SGD \\  NH = 512, NL = 3 \end{tabular} \\ 
 SI & \begin{tabular}{@{}c@{}}lr = 1e-3, SGD \\  NH = 512, NL = 3 \end{tabular} & \begin{tabular}{@{}c@{}}lr = 1e-3, SGD+M \\  NH = 512, NL = 3 \end{tabular} & \begin{tabular}{@{}c@{}}lr = 1e-3, SGD \\  NH = 512, NL = 3 \end{tabular} \\ 
 MAS & \begin{tabular}{@{}c@{}}lr = 1e-4, Adam \\  NH = 512, NL = 3 \end{tabular} & \begin{tabular}{@{}c@{}}lr = 1e-4, Adam \\  NH = 512, NL = 3 \end{tabular} & \begin{tabular}{@{}c@{}}lr = 2e-4, Adam \\  NH = 512, NL = 3 \end{tabular} \\ 
 Lwf & \begin{tabular}{@{}c@{}}lr = 1e-3, SGD \\  NH = 1024, NL = 2 \end{tabular} & \begin{tabular}{@{}c@{}}lr = 1e-3, SGD \\  NH = 1024, NL = 2 \end{tabular} & \begin{tabular}{@{}c@{}}lr = 1e-3, SGD \\  NH = 1024, NL = 2 \end{tabular} \\  
 ICarl & \begin{tabular}{@{}c@{}}lr = 1e-3, SGD+M \\  NH = 1024, NL = 2 \end{tabular} & \begin{tabular}{@{}c@{}}lr = 1e-3, SGD+M \\  NH = 1000, NL = 2 \end{tabular} & \begin{tabular}{@{}c@{}}lr = 1e-4, SGD+M \\  NH = 1000, NL = 2 \end{tabular} \\ 
 Lucir & \begin{tabular}{@{}c@{}}lr = 1e-4, Adam \\  NH = 512, NL = 3 \end{tabular} & \begin{tabular}{@{}c@{}}lr = 2e-4, AdamW \\  NH = 512, NL = 3 \end{tabular} & \begin{tabular}{@{}c@{}}lr = 2e-5, AdamW \\  NH = 512, NL = 3 \end{tabular} \\ 
 GDumb & \begin{tabular}{@{}c@{}}lr = 2e-4, AdamW \\  NH = 512, NL = 3 \end{tabular} & \begin{tabular}{@{}c@{}}lr = 2e-5, Adam \\  NH = 512, NL = 3 \end{tabular} & \begin{tabular}{@{}c@{}}lr = 2e-4, AdamW \\  NH =512 , NL = 3 \end{tabular} \\  
 Mnem & \begin{tabular}{@{}c@{}}lr = 1e-4, Adam \\  NH = 512, NL = 3 \end{tabular} & \begin{tabular}{@{}c@{}}lr = 2e-5, Adam \\  NH = 500, NL = 3 \end{tabular} & \begin{tabular}{@{}c@{}}lr = 2e-4, Adam \\  NH = 512, NL = 3 \end{tabular} \\ 
 S-NCN & \begin{tabular}{@{}c@{}}lr = 0.0105, SGD \\  NH = $500$, NL = $3$ \end{tabular} & \begin{tabular}{@{}c@{}}lr = 0.01, SGD \\  NH = $500$, NL = $3$ \end{tabular} & \begin{tabular}{@{}c@{}}lr = 0.011, SGD \\  NH = $500$, NL = $3$ \end{tabular} \\ 
 \hline
\end{tabular}
}
\end{center}
    \caption{\textcolor{black}{
    General hyper-parameter values selected from tuning. Models above the double horizontal line are multi-head models and models below it are single-head models. ``NH'' stands for ``number of hidden neurons'', ``NL'' stands for ``number of hidden layers'', ``lr'' stands for ``learning rate'', and optimizer choice is either ``SGD'' for ``stochastic gradient descent'' (``SGA'' means ``stochastic gradient ascent''), ``Adam'', ``AdamW'', or ``SGD+M'' for ``SGD with momentum''.
    }}
    \label{tab:baseline_hyperparameters}
\end{table}

\section*{Discussion: On the Limitations of Sequential Neural Coding}
\label{sec:limitations}

Our model jointly predicts the target label and learns to generate the sensory input, further driven/modulated by a simple complementary neural system that mitigates neural cross-talk. The dual nature of our model/system helps to uncover distributed representations that facilitate robust learning and adaptation over sequences of tasks/datasets. Even though this design scheme provides flexibility and seems to offer many advantages compared to other backprop-based models, it does come with several limitations. Mainly, finding the true posterior distribution over latent neural activities is harder than just learning a forward mapping between inputs and output targets and, notably, it can be expensive to find a good set of neural activity values as the problem complexity increases (notably the $K$ hyper-parameter, which controls the amount of steps taken per data point/mini-batch by the S-NCN to iteratively infer a potential maximum a posterior estimate of its state variables). Currently, the S-NCN conducts inference and learning through a sort of expectation-maximization process and, fortunately, in the problems we studied, the value of $K$ was fairly low (only $10$ to $20$ steps at most were needed to find useful state values per sample/mini-batch). However, for more complex data types, such as natural images with multiple objects and even background scenery, the value of $K$ will quite likely need to be much higher, increasing the computation time further needed to conduct online inference. This drawback could be mitigated by potentially integrating mechanisms to support amortized inference, e.g., predictive sparse decomposition \cite{kavukcuoglu2010fast}, and by designing custom software/hardware implementations that exploit the actual layer-wise parallelism (which could work in asynchronous settings) that the S-NCN model offers for both inference and weight updating.

Furthermore, the fact that the S-NCN (in its current form/implementation in this study) must solve a dual optimization problem that entails jointly learning to predict the target label and generate the sensory input (image) might compromise the model's overall accuracy when tested on large-scale images. It is often an easier problem to directly learn a conditional mapping between the input and label as opposed to learning a full generative model as the S-NCN does \cite{ng2002discriminative}. Future work should explore adapting the S-NCN to only learn a conditional mapping as opposed to a full joint distribution over inputs and labels as well as potential mechanisms for pre-training the generative side of the system (which would allow freezing of the synaptic weights for generation and only require updating discriminative problems -- this could potentially reduce the value of $K$ even for more complex sensory inputs). Another drawback, yet also simultaneously a strength, is the fact that the S-NCN is attempting to optimize (online) total discrepancy as opposed to a single, global surface loss. While total discrepancy is one important key to breaking free of backprop and its limitations, i.e., it naturally facilitates a local learning problem without the need for a global feedback pathway, it also creates a more challenging optimization problem in general, i.e., the neural system must now not only match the values created by data but also ensure its internal activities and its local predictions of each of them are aligned. While the overall complementary system largely mitigates catastrophic interference (or the neural cross-talk that would trigger the loss/deletion of previously acquired knowledge), this primarily affects measurements of backward transfer (BWT) but could potentially damage the model's per-task performance, i.e., the main diagonal of the task matrix. Since we do not impose any strong distributional assumptions over the latent activities (such as a clean Gaussian prior as is often done in variational autoencoders), if the S-NCN's estimated value of the latent activities are far from the true posterior, then the S-NCN might produce sub-optimal performance, especially for complex problems.
Even though all continual learning systems suffer from this issue (especially most modern-day continual learning ANNs), our model's dual optimization nature could experience this problem more frequently. We believe that integrating memory-aware retrieval from synapses, a mechanism guided by (a brain-like form of) replay, can serve as a plausible solution moving forward.  This might help the system by directing it to be closer to the true posterior by avoiding bad local minima when learning continuously. 

Additionally, with respect to our proposed task selection mechanism \textcolor{black}{(the functional neural basal ganglia circuit)}, one notable drawback is that a small refractory period is imposed in order to ensure that enough data is accumulated from the stream to update the competitive task selector's weights. This would be an issue for task streams that constantly introduce tasks with fewer samples than the pre-set refractory period. A subject of our future work is to improve the power /adaptability of the task selection model in the face of more volatile task sequence streams. 
\textcolor{black}{
Another limitation is that the S-NCN is, in effect, a dynamically-expanding architecture: there is an overhead for the task-context memory – one new task context vector would need to be generated/grown for each new (disjoint) task is encountered. While this required parameter growth/generation is not as high as other dynamically-expanding architecture approaches (such as progressive networks \cite{rusu2016progressive} or dynamically-expanding networks \cite{yoon2017lifelong}, where many new parameters must be created per task). While the inclusion of relatively few, additional context vector parameters is more desirable, requiring the growing out of parameters at a rate far less than methods such as \cite{rusu2016progressive,yoon2017lifelong}. 
To mitigate the cost that even the S-NCN imposes, we remark that the S-NCN’s generative cortex could be adapted to induce its own form of layer-wise replay as one alternative, similar to \cite{van2020brain}, or that another circuit could be designed to potentially learn how to compress these task contexts by reducing redundancy exploiting overlap/redundancy between contexts (serving as a form of efficient long-term memory).}

Finally, a more obvious drawback is that the S-NCN's error synapses also increase the memory footprint of the overall model.
It would be advisable, when using/applying a model like the S-NCN on other continual learning problems, to select the number of hidden layers and number of neurons in each layer based on model capacity, i.e., compute the total number of (generative and error) synapses that would result from making the neural structure more complex or deeper. A more long-term, promising means of mitigating the increased memory footprint would be to design error units further inspired by actual neurons -- instead of assuming a one-to-one mapping (one error neuron per state neuron), design small pools of neurons that are responsible for computing the mismatch activities for large groups of state neurons. This is a key solution to investigate in future work.

\section*{Creating Task Orderings \#1 and \#2}

%

To create our sequential learning benchmarks, we utilize the MNIST, Fashion MNIST, and Google Draw datasets to create various sets of ``subtasks'', or rather, classification problems that involve different classes of the original set of each full dataset. In this paper, we create a 6 task sequence, $\{T_1,T_2,T_3,T_4,T_5,T_6 \}$, from these datasets, where two tasks are generated from each specific dataset. To create the task splits, we create data subsets based on minimizing the amount of knowledge transfer across data splits, specifically by examining the amount of stroke overlap in the images across classes, yielding a challenging problem. For equal number of classes, the splits we created were: MNIST set \#1, \emph{M1} = \{$0,8,3,5,2$\}, MNIST set \#2, \emph{M2} = \{$1,4,6,7,9$\}, Fashion MNIST set \#1, \emph{FM1} = \{top, trouser, pullover, dress, coat\}, Fashion MNIST set \#2, \emph{FM2} = \{sandal, shirt, sneaker, bag, ankle boot\}, Google Draw set \#1, \emph{GD1} = \{objects that were car or bike variants \}, and Google Draw set \#2, \emph{GD2} = \{objects belong to variants of airplanes or submarines \}.
For a task sequence, we create two scenarios: 1) where number of classes are equal for all tasks (i.e., 5 classes in our setup), and 2) where number of classes are unequal (number of classes per task was chosen randomly, omitting the number $5$ as an option). 
In our experiments, we investigate two task orderings (Ordering \#1 and Ordering \#2). We compute the color index similarity \cite{swain1991color} between every pair of tasks (as a proxy for task similarity) and randomly chose Orderings \# 1 and \# 2 so that the color similarity between adjacent tasks was higher for Ordering \#1 (``High Color Sim.'' for high color similarity) than for Ordering \#2 (``Low Color Sim.'' for low color similarity), hence task transfer should be easier for Ordering \#1 than \#2. 
These particular orderings could be considered to be ``harder`` and ``easier'' task orderings, respectively, since it is possible that a the difference in color-index would make it easier to differentiate the tasks (the diversity of inputs from the first few tasks might even improve the performance, as it would be in the case of Ordering \# 2). We can see this reflected by the fact that all models (the S-NCN and the baselines) perform a bit better in general on Ordering \# 2 (low color similarity or ``easier'' ordering) than on Ordering \# 1 (high color similarity or ``harder'' ordering).

%
%

\section*{Expanded Results for Task Orderings \# 1 and \# 2}

\section*{Metrics for Quantifying Memory Retention}

The formulas for ACC and BWT are:
\begin{align}
\mbox{\normalsize ACC} = \frac{1}{T}
\sum_{i=1}^T R_{T,i},\quad
\mbox{\normalsize BWT} = \frac{1}{T-1}
\sum_{i=1}^{T-1} R_{T,i} - R_{i,i}  \mbox{.} \label{eq:acc_and_bwt}
\end{align}

In addition, \textbf{we propose two additional, complementary metrics}, with the motivation that these metrics examine aspects of forgetting and capacity not clearly captured by ACC or BWT. Our two measures, True BWT (TBWT) and Cumulative BWT for task $T_t$ (CBWT(t)), are defined as follows:
\begin{align}
\mbox{\normalsize CBWT(t)} &= \frac{1}{T-t}\sum_{i=t+1}^T R_{i,t} - R_{t,t} \\
\mbox{\normalsize TBWT} &= \frac{1}{T-1}
\sum_{i=1}^{T-1} R_{T,i} - G_{i,i} \label{eqn:cbwt_and_tbwt}
\end{align}
where $G_{i,i}$ is the performance of an independent classifier trained on task $i$ (in our experiments, this was a full capacity MLP trained via backprop). TBWT relates the degradation in prior task performance by replacing the diagonal of task matrix $R$ with a ``gold standard'', which is the performance of a model that, in isolation, is able to allocate its full capacity to a particular task. CBWT(t) is a task-specific metric, where we instead examine a particular column of $R$, and measure the total amount of forgetting throughout the sequential learning process, instead of simply examining the final performance at the end (bottom row of $R$) as BWT can only do. CBWT(t) would punish models that suffer large dips in performance in the middle of learning (but not necessarily at the end), and would be better suited for characterizing forgetting in stream settings than BWT.

\begin{sidewaystable}
\begin{center}
\caption{Alternative metrics reported for task sequence orderings \#1 and \#2 (higher values are better).}
\label{results:alternative_metrics}
\begin{tabular}{|l||c|c||c|c||c|c||c|c|}
\hline
  \multicolumn{1}{|l||}{}&\multicolumn{4}{c||}{\begin{tabular}[x]{@{}c@{}}\textbf{Ordering \#1 (High Color Sim.)}\\\end{tabular}}&\multicolumn{4}{c|}{\begin{tabular}[x]{@{}c@{}}\textbf{Ordering \#2 (Low Color Sim.)}\\\end{tabular}} \tabularnewline
  \multicolumn{1}{|l||}{}&\multicolumn{2}{c}{\begin{tabular}[x]{@{}c@{}}\textbf{Equal}\\\end{tabular}}&\multicolumn{2}{c||}{\begin{tabular}[x]{@{}c@{}}\textbf{Unequal}\\\end{tabular}}&\multicolumn{2}{c}{\begin{tabular}[x]{@{}c@{}}\textbf{Equal}\\\end{tabular}}&\multicolumn{2}{c|}{\begin{tabular}[x]{@{}c@{}}\textbf{Unequal}\\\end{tabular}}\tabularnewline
  & \textbf{TBWT} & \textbf{CBWT} & \textbf{TBWT} & \textbf{CBWT} & \textbf{TBWT} & \textbf{CBWT} & \textbf{TBWT} & \textbf{CBWT} \\
  \hline
  Backprop & -0.426 & -0.358 & -0.547 & -0.496 & -0.422 & -0.566 & -0.602 & -0.639 \\
  EWC &  -0.409 &  -0.332&  -0.516 &  0.477 & -0.400 &  -0.521 &  -0.599 &  -0.611 \\
  Md-IMM &  -0.388 &  -0.296 &  -0.481 &  -0.390 &  -0.355 &  -0.411 & -0.521 & -0.429 \\
  DT+Md-IMM & -0.342 & -0.281 & -0.466 & -0.355 & -0.340 & -0.399 & -0.491 & -0.389 \\
  L2+DT+Md-IMM & -0.301 & -0.250 & -0.422 & -0.318 & -0.281 & -0.355 & -0.401 & -0.378 \\
  HAT & -0.277 & -0.252 & -0.341 & -0.291 & -0.200 & -0.341& -0.285 & -0.328 \\
 \hline
  S-NCN & $-0.397$ & $-0.509$ & $-0.507$ & $-0.623$ & $-0.373$ & $-0.252$ & $-0.564$ & $-0.396$ \\
  Lat1-S-NCN & $\mathbf{-0.048}$ & $\mathbf{-0.074}$ & $\mathbf{-0.081}$ & $\mathbf{-0.048}$ & $\mathbf{-0.032}$ & $\mathbf{-0.150}$ & $\mathbf{-0.086}$ & $\mathbf{-0.037}$ \\
  Lat2-S-NCN & $-0.226$ & $-0.304$ & $-0.270$ & $-0.310$ & $-0.145$ & $-0.215$ & $-0.198$ & $-0.273$ \\ 
  \hline
\end{tabular}
\end{center}
\end{sidewaystable}

\begin{table*}[!t]
\begin{center}
\caption{Generalization metrics (10 trials) for sequence orderings \# 1 \& \#2 (higher values are better).}
\label{appendix_results:seq_variant}
\footnotesize
\begin{tabular}{|l||c|c||c|c|}
\hline
  \multicolumn{1}{|l||}{}&\multicolumn{4}{c|}{\begin{tabular}[x]{@{}c@{}}\textbf{Ordering \#1: $\{M1, M2, GD1, FM1, FM2, GD2\}$ (High Color Sim.)}\\\end{tabular}} \tabularnewline
  \multicolumn{1}{|l||}{}&\multicolumn{2}{c}{\begin{tabular}[x]{@{}c@{}}\textbf{Equal}\\\end{tabular}}&\multicolumn{2}{c|}{\begin{tabular}[x]{@{}c@{}}\textbf{Unequal}\\\end{tabular}}\tabularnewline
  & \textbf{ACC} & \textbf{BWT} & \textbf{ACC} & \textbf{BWT} \\
  \hline
  Backprop & $0.241\pm0.050$ & $-0.759\pm0.030$ & $0.185\pm0.048$ & $-0.791\pm0.048$ \\
  Backprop+DO & $0.251\pm0.049$ & $-0.711\pm0.030$ & $0.178\pm0.049$ & $-0.733\pm0.049$ \\
  EWC & $0.280\pm0.023$ & $-0.714\pm0.030$ & $0.185\pm0.046$ & $-0.726\pm0.039$ \\
  EWC+DO & $0.231\pm0.029$ & $-0.687\pm0.029$ & $0.184\pm0.044$ & $-0.710\pm0.038$ \\
  Mean-IMM & $0.279\pm0.019$ & $-0.465\pm0.024$ & $0.210\pm0.041$ & $-0.499\pm0.043$ \\
  Md-IMM & $0.521\pm0.027$ & $-0.392\pm0.023$ & $0.480\pm0.039$ & $-0.240\pm0.040$ \\
  DT+Mean-IMM & $0.321\pm0.023$ & $-0.430\pm0.020$ & $0.300\pm0.044$ & $-0.471\pm0.044$ \\
  DT+Md-IMM & $0.530\pm0.024$ & $-0.387\pm0.021$ & $0.551\pm0.042$ & $-0.220\pm0.042$ \\
  L2+Mean-IMM & $0.301\pm0.022$ & $-0.443\pm0.022$ & $0.250\pm0.038$ & $-0.492\pm0.046$ \\
  L2+Md-IMM & $0.491\pm0.020$ & $-0.376\pm0.023$ & $0.480\pm0.039$ & $-0.235\pm0.041$ \\
  L2+DT+Mean-IMM & $0.354\pm0.029$ & $-0.390\pm0.021$ & $0.351\pm0.039$ & $-0.421\pm0.046$ \\
  L2+DT+Md-IMM & $0.532\pm0.025$ & $-0.237\pm0.027$ & $0.520\pm0.040$ & $-0.240\pm0.045$ \\
  HAT & $0.550\pm0.019$ & $-0.211\pm0.020$ & $0.492\pm0.031$ & $-0.231\pm0.036$ \\
  \hline
  S-NCN (ours)  & $0.421 \pm 0.022$ & $-0.408 \pm 0.026$ & $0.352 \pm 0.016$ & $-0.476 \pm 0.020$ \\
  S-NCN-relu (ours)  & $0.398 \pm 0.009$ & $-0.430 \pm 0.012$ & $0.352 \pm 0.008$ & $-0.470 \pm 0.011$ \\
  Lat1-S-NCN (ours) & $\mathbf{0.716 \pm 0.013}$ & $\mathbf{-0.031 \pm 0.017}$ & $\mathbf{0.713 \pm 0.011}$ & $\mathbf{-0.041 \pm 0.012}$ \\
  Lat2-S-NCN (ours) & $0.573 \pm 0.020$ & $-0.236 \pm 0.0258$ & $0.554 \pm 0.038$ & $ -0.235 \pm 0.042$ \\ 
  \hline
  \hline 
  \multicolumn{1}{|l||}{}&\multicolumn{4}{c|}{\begin{tabular}[x]{@{}c@{}}\textbf{Ordering \#2: $\{GD2, M1, FM2, M2, GD1, FM1\}$ (Low Color Sim.)}\\\end{tabular}} \tabularnewline
  & \textbf{ACC} & \textbf{BWT} & \textbf{ACC} & \textbf{BWT} \\
  \hline
  Backprop & $0.303\pm0.030$ & $-0.644\pm0.037$ & $0.287\pm0.043$ & $-0.671\pm0.043$ \\
  Backprop+DO & $0.285\pm0.032$ & $-0.587\pm0.031$ & $0.266\pm0.042$ & $-0.610\pm0.044$ \\
  EWC & $0.303\pm0.031$ & $-0.643\pm0.033$ & $0.291\pm0.039$ & $-0.663\pm0.047$ \\
  EWC+DO & $0.302\pm0.033$ & $-0.558\pm0.032$ & $0.281\pm0.039$ & $-0.586\pm0.046$ \\
  Mean-IMM & $0.453\pm0.026$ & $-0.170\pm0.031$ & $0.402\pm0.036$ & $-0.274\pm0.035$ \\
  Md-IMM & $0.584\pm0.027$ & $-0.091\pm0.030$ & $0.533\pm0.034$ & $-0.230\pm0.036$ \\
  DT+Mean-IMM & $0.558\pm0.021$ & $-0.128\pm0.029$ & $0.510\pm0.033$ & $-0.254\pm0.035$ \\
  DT+Md-IMM & $0.591\pm0.020$ & $-0.088\pm0.032$ & $0.528\pm0.036$ & $-0.211\pm0.039$ \\
  L2+Mean-IMM & $0.465\pm0.021$ & $-0.156\pm0.033$ & $0.430\pm0.039$ & $-0.271\pm0.032$ \\
  L2+Md-IMM & $0.576\pm0.028$ & $-0.99\pm0.038$ & $0.511\pm0.036$ & $-0.266\pm0.039$ \\
  L2+DT+Mean-IMM & $0.587\pm0.025$ & $-0.105\pm0.033$ & $0.528\pm0.038$ & $-0.253\pm0.043$ \\
  L2+DT+Md-IMM & $0.630\pm0.029$ & $-0.076\pm0.030$ & $0.551\pm0.037$ & $-0.201\pm0.041$ \\
  HAT & $0.596\pm0.026$ & $-0.114\pm0.029$ & $0.563\pm0.031$ & $-0.210\pm0.044$ \\
  \hline
  S-NCN (ours)  & $0.444 \pm 0.017$ & $-0.393 \pm 0.0210$ & $0.272 \pm 0.013$ & $-0.587 \pm 0.014$ \\
  S-NCN-relu (ours)  & $ 0.431 \pm 0.009$ & $-0.398 \pm 0.010$ & $0.286 \pm 0.007$ & $-0.559 \pm 0.008$ \\
  Lat1-S-NCN (ours) & $\mathbf{0.721 \pm 0.014}$ & $\mathbf{-0.042 \pm 0.013}$ & $\mathbf{0.667 \pm 0.011}$ & $\mathbf{-0.097 \pm 0.013}$ \\
  Lat2-S-NCN (ours) & $0.633 \pm 0.028$ & $-0.170 \pm 0.033$ & $0.5778 \pm 0.035$ & $-0.211 \pm 0.042$ \\
  \hline
\end{tabular}
\end{center}
\end{table*}

\begin{sidewaystable}
\centering
\caption{Generalization metrics (10 trials) for Split MNIST, Split Fashion MNIST (FMNIST) and Not-MNIST benchmarks. Note for IMM, we employ the best performing variant, \emph{L2+DT+Md-IMM}. (SH indicates single-head while MH indicates multi-head.)
}
\label{appendix_results:benchmarks}
\begin{tabular}{|l||c|c||c|c|c|c|}
\hline
  \multicolumn{1}{|l||}{}&\multicolumn{2}{c}{\begin{tabular}[x]{@{}c@{}}\textbf{MNIST}\\\end{tabular}}&\multicolumn{2}{c|}{\begin{tabular}[x]{@{}c@{}}\textbf{FMNIST}\\\end{tabular}}&\multicolumn{2}{c|}{\begin{tabular}[x]{@{}c@{}}\textbf{NotMNIST}\\\end{tabular}}\tabularnewline
  & \textbf{ACC} & \textbf{BWT} & \textbf{ACC} & \textbf{BWT} & \textbf{ACC} & \textbf{BWT}\\
  \hline
  \hline
  EWC (MH) & $0.760 \pm 0.030$ & $-0.210 \pm 0.011$ & $0.739 \pm 0.020$ & $-0.201 \pm 0.011$ & $0.790 \pm 0.020$ & $-0.176 \pm 0.010$ \\
  VCL (MH) & $0.980 \pm 0.210$ & $-0.003 \pm 0.002$  & $0.980 \pm 0.20$ & $-0.002 \pm 0.003$ & $0.953 \pm 0.003$ & $-0.004 \pm 0.002$\\
  IMM (MH) & $0.951 \pm 0.018$ & $-0.007 \pm 0.003$ & $0.950 \pm 0.013$ & $-0.005 \pm 0.003$ & $0.925 \pm 0.011$ & $-0.006 \pm 0.002$ \\
  HAT (MH) & $0.972 \pm 0.011$ & $-0.040 \pm 0.002$  & $0.968 \pm 0.011$ & $-0.004 \pm 0.002$ & $0.942 \pm 0.009$ & $-0.005 \pm 0.002$  \\
  GEM (MH) & $0.922 \pm 0.110$ & $\mathbf{+0.001 \pm 0.002}$  & $0.930 \pm 0.12$ & $+0.001 \pm 0.003$ & $0.919 \pm 0.021$ & $-0.003 \pm 0.002$  \\
  DGR (MH) & $0.911 \pm 0.300$ & $-0.011 \pm 0.002$ & $0.915 \pm 0.25$ & $-0.013 \pm 0.001$ & $0.920 \pm 0.015$ & $-0.014 \pm 0.003$  \\
  Rtf (MH) & $0.925 \pm 0.200$ & $-0.009 \pm 0.003$  & $0.930 \pm 0.25$& $-0.009 \pm 0.005$ & $0.922 \pm 0.012$ & $-0.011 \pm 0.004$ \\
  \hdashline
  EWC (SH) & $0.190 \pm 0.030$ & $-0.357 \pm 0.015$ & $0.199 \pm 0.06$ & $-0.350 \pm 0.012$ & $0.186 \pm 0.020$ & $-0.361 \pm 0.010$  \\
  NR+M1 (SH) & $0.906 \pm 0.870$ & $-0.050 \pm 0.001$  & $0.900 \pm 0.810$& $-0.060 \pm 0.003$ & $0.890 \pm 0.030$ & $-0.071 \pm 0.004$   \\
  NR+M2 (SH) & $0.950 \pm 0.470$ & $-0.100 \pm 0.003$  & $0.948 \pm 0.380$ & $-0.090 \pm 0.003$ & $0.880 \pm 0.028$ & $-0.103 \pm 0.002$   \\
  SI (SH) & $0.197 \pm 0.110$ & $-0.367 \pm 0.014$  & $0.198 \pm 0.10$ & $-0.370 \pm 0.013$ & $0.161 \pm 0.030$ & $-0.370 \pm 0.010$  \\
  MAS (SH) & $0.195 \pm 0.290$ & $-0.340 \pm 0.010$  & $0.180 \pm 0.25$  & $-0.340 \pm 0.010$ & $0.178 \pm 0.060$ & $-0.341 \pm 0.011$  \\
  Lwf (SH) & $0.846 \pm 0.340$ & $-0.120 \pm 0.001$  & $0.875 \pm 0.30$ & $-0.130 \pm 0.003$ & $0.626 \pm 0.091$ & $-0.130 \pm 0.004$ \\
  ICarl (SH) & $0.940 \pm 0.410$ & $-0.100 \pm 0.004$  & $0.960 \pm 0.40$ & $-0.110 \pm 0.005$ & $0.887 \pm 0.102$ & $-0.109 \pm 0.007$  \\
  Lucir (SH) & $0.940 \pm 0.310$ & $-0.103 \pm 0.007$   & $0.950 \pm 0.34$ & $-0.110 \pm 0.005$ & $0.935 \pm 0.093$ & $-0.101 \pm 0.006$  \\
  Bic (SH) & $0.901 \pm 0.860$ & $-0.139 \pm 0.009$  & $0.890 \pm 0.85$ & $-0.160 \pm 0.009$ & $0.851 \pm 0.099$ & $-0.155 \pm 0.009$ \\
  Mnem (SH)  & $0.960 \pm 0.320$ & $-0.991 \pm 0.005$  & $0.968 \pm 0.30$ & $\mathbf{+0.007 \pm 0.006}$ & $0.950 \pm 0.071$ & $-0.011 \pm 0.007$  \\
  \hline
  S-NCN (SH) & $\mathbf{0.981 \pm 0.300}$ & $-0.005 \pm 0.004$   & $\mathbf{0.982 \pm 0.400}$ & $-0.003 \pm 0.007$ & $\mathbf{0.957 \pm 0.400}$ & $\mathbf{-0.004 \pm 0.005}$ \\
  \hline
\end{tabular}
\end{sidewaystable}

\paragraph{Discussion: }
Results are reported in Tables \ref{results:alternative_metrics} and  \ref{appendix_results:seq_variant} (an expanded version of the original one in the main paper). Each simulation was run $10$ times, each trial using a unique seed for pseudo-random number generation, we report both mean and standard deviation.
As we observe in our experimental results, we see that all of our S-NCN models exhibit improved memory retention over simple baselines, such as backprop, and more notably, EWC. However, we see that incorporating task-driven lateral inhibition in facilitating gradual forgetting as opposed to catastrophic forgetting, as evidenced by the very competitive performance of both Lat1-S-NCN and Lat2-S-NCNs, with Lat1-S-NCN outperforming all baselines consistently, in terms of both ACC and BWT. This result is robust across both task sequences and equal/ unequal class settings. It is further important to note that the meta-parameter settings used for the various S-NCNs were only tweaked minorly with the same values across all of the settings/scenarios. The observation that lateral inhibition improves the neural computation of our interactive network further corroborates the result of \cite{oreilly2001inhibition}, though it focused on models trained via contrastive Hebbian learning.

Upon examination of Table \ref{results:alternative_metrics}, in terms of TBWT and CBWT(1)\footnote{We measure CBWT for task $T_1$, since this measures total forgetting over the full length of the task sequence.}, the proposed lateral S-NCNs still outperform the baselines. The key difference is that we see that the lateral S-NCNs actually do retain prior information throughout learning and do not simply just recover it at the end.

\section*{Expanded Benchmark Results}

To start, we describe the three forms of the lateral competition we designed for the S-NCN. They were specifically as follows: 
\begin{enumerate}
    \item $f^\ell(\mathbf{z}^\ell,\mathbf{g}^\ell_t) = \mathbf{I} \cdot \mathbf{z}^\ell$, which means that the lateral inhibitory matrix is fixed to a diagonal matrix $\mathbf{I}$ and forces the model to ignore the task embedding (in the Appendix, we denote this as ``NoLat-S-NCN''),
    \item $f^\ell(\mathbf{z}^\ell, \mathbf{g}_t^\ell) = \big( \mathbf{I} \odot \mathbf{V}^\ell \big) \cdot \mathbf{z}^\ell $, where the matrix
    $\mathbf{V}^\ell = \Call{bkWTA}{\mathbf{g}^\ell_t,K} \cdot \Call{bkWTA}{(\mathbf{g}^\ell_t)^T,K}$ and
    $\Call{bkWTA}{\mathbf{v},K}$ is the binarized $K$ winners-take-all function, yielding a binary vector with $1$ at the index of each of the $K$ winning units (in the Appendix, we denote this as ``Lat1-S-NCN''), 
    \item $f^\ell(\mathbf{z}^\ell, \mathbf{g}_t^\ell) = \max \big(0,\mathbf{z}^\ell - \big( (\mathbf{V}^\ell \odot ( \mathbf{g}^\ell_t \cdot (\mathbf{g}^\ell_t)^T) ) \cdot \mathbf{z}^\ell \big) \big)$, where: $\mathbf{V}^\ell_{i,j} = \{\alpha \mbox{, if } i \neq j \mbox{, else, } 0 \}$  (in the Appendix, we denote this as ``Lat2-S-NCN'').
\end{enumerate}
In the last two forms of the competition function, we see that lateral inhibition is a function of an evolving context vector $\mathbf{g}^\ell$, triggered by the presence of the task signal/pointer $t_i$. 
The above competition functions correspond to different designs of lateral suppression patterns: 
(1) corresponds to no lateral suppression,
(2) corresponds to shutting off neurons that are not task-relevant driven by a task selector, 
(3) corresponds to a task-driven, real-valued lateral matrix that scales the neural activities. 

On the custom benchmarks (including both orderings \#1 and \#2), we evaluate four variations of the S-NCN: 1) an S-NCN, with hyperbolic tangent activations and no context-dependent lateral inhibition (S-NCN), 2) an S-NCN with sparsity created by the use of a linear rectifier activation function and no lateral inhibition (S-NCN-relu), 3) an S-NCN with the second variant of our proposed lateral inhibition (Lat1-S-NCN), and 4) an S-NCN with the third variant of our proposed lateral inhibition (Lat2-S-NCN) (All variants used: $\beta = 0.05$, $K = 10$, $\eta_g = 0.9$, $\eta_e = 0.01$, $\alpha = 0.98$). The last two S-NCN models (``Lat1-S-NCN'' and ``Lat2-S-NCN'') were driven by the FNBG model that we described in Section 3.3 (``The Neural Task Selection Model''). 

We compare our S-NCN model (specifically, our best-performing one from the experiment in the last section -- the Lat1-S-NCN) to the following approaches that have been proposed over the years to combat catastrophic interference: 
Naive rehearsal with memory (NR+M1 \& NR+M2), EWC, 
synaptic intelligence (SI), MAS \cite{mas}, Lwf \cite{lwf}, GEM \cite{gem}, DGR \cite{dgr}, Rtf \cite{van2019three}, ICarl \cite{icarl}, Lucir \cite{lucir}, Bic \cite{bic}, and Mnemonics \cite{mnemonics}. In Table \ref{appendix_results:benchmarks}, we report model ACC and BWT, averaged over $10$ trials, offering not only an extensive and comprehensive comparison of competitive methods, but also demonstrating that, for all three benchmarks, \textbf{our proposed S-NCN outperforms all of them}, demonstrating the power afforded by challenging the very assumptions underlying modern-day artificial neural systems and designing models with stronger grounding in neuro-biology. Furthermore, it is astounding to see that the S-NCN outperforms/matches performance with not only the single-head models but also with the multi-head models (except GEM), which enjoy an easier version of the problem since they are permitted to utilize a different classifier per task. Finally, it is critical to note that \textbf{our proposed S-NCN is a complementary neural system that learns without explicitly-provided task descriptors}, i.e., in other words, the model learns to compose its own task contexts in a data-dependent manner.

%
Note that, in the online setting, split FashionMNIST appears to be simpler, given that the S-NCN readily learns to generate rough forms of bigger objects (shirt/shoes/pants) and associate these with labels early whereas digits/characters are a bit more intricate and take longer to learn.

\subsection*{\textcolor{black}{S-NCN Task Accuracy Curves}}

\begin{figure}[!t]
 \centering
 \includegraphics[width=0.9\textwidth]{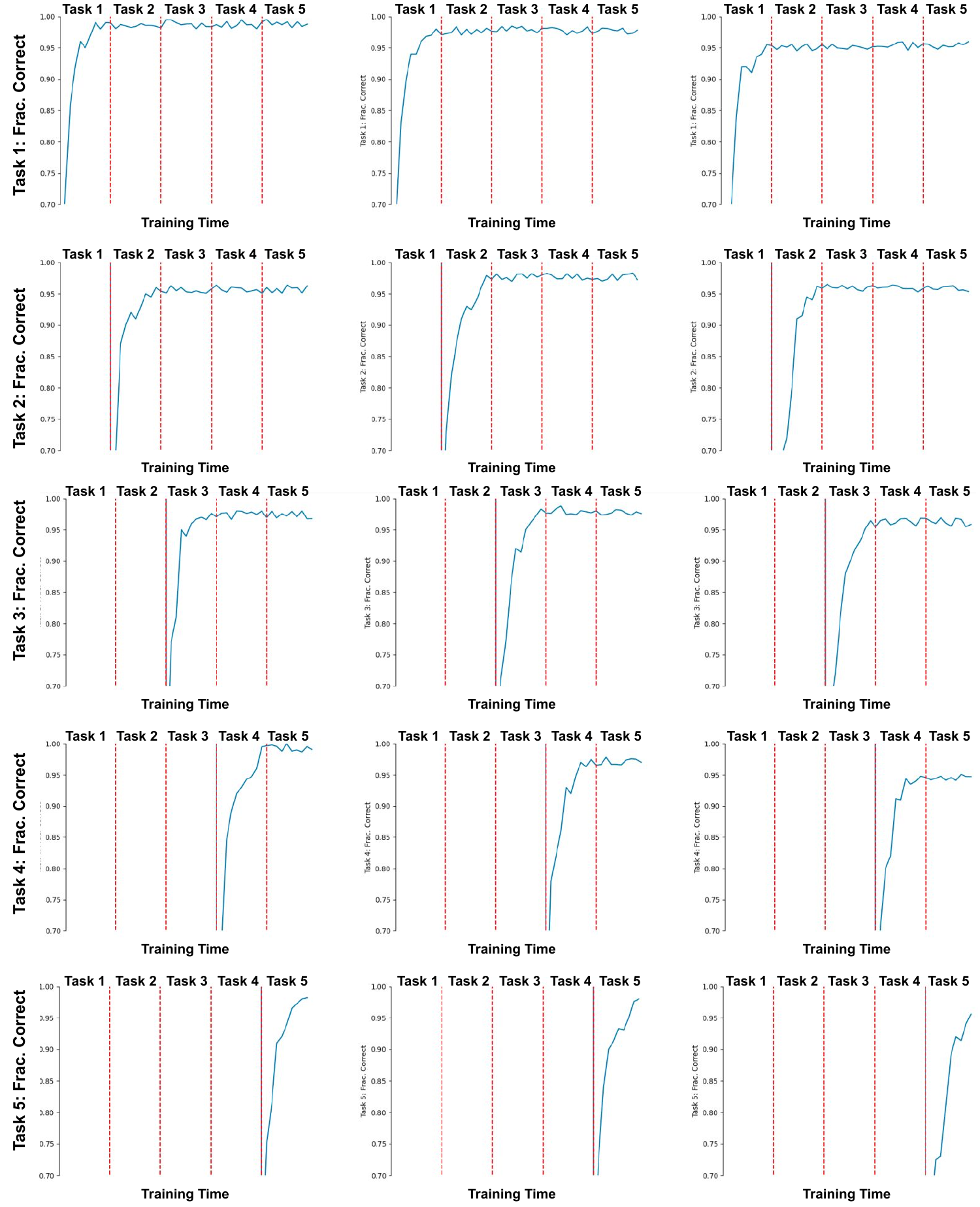}
\vspace{-0.3cm}
\caption{\textcolor{black}{Task accuracy curves for the S-NCN measured across the five tasks within the (Left column) Split MNIST, (Middle column) Split Fashion MNIST, and (Right column) Split NotMNIST benchmarks (red vertical lines indicate actual task boundaries). Y-axis depicts the fraction correct while the x-axis depicts (online) training iteration (or one single epoch/pass through each task). Each row, as indicated by the Y-axis label, represents the perspective of a different task, e.g., row 1 corresponds to performance on Task 1 as the S-NCN learns across all tasks, row 2 corresponds to performance on Task 2 as the S-NCN learns across all tasks, etc.}}
\label{fig:task_acc_curves}
\vspace{-0.5cm}
\end{figure}

\textcolor{black}{
See Figure \ref{fig:task_acc_curves} for a visual depiction of the S-NCN's task accuracy over time across tasks for all three continual learning benchmarks investigated in this paper.
}

